\documentclass[10pt,twocolumn,letterpaper]{article}

\usepackage{cvpr}              %

\usepackage[dvipsnames]{xcolor}
\usepackage{soul}
\usepackage{svg}
\usepackage[misc]{ifsym}
\usepackage{enumitem}
\newcommand{\paperpar}[1]{\vspace{.10cm}\noindent{\bf #1:}}
\usepackage{comment}
\usepackage{float}

\usepackage{amsmath}
\usepackage{amssymb}
\usepackage{booktabs}
\usepackage{bm}
\usepackage{cancel}
\usepackage{math_commands}

\usepackage{graphicx}
\usepackage{pifont}%
\usepackage{tikz}
\usetikzlibrary{positioning}
\usetikzlibrary{shapes.geometric}

\usepackage[noend,ruled,linesnumbered]{algorithm2e}
\SetKwInput{KwInput}{Input}                %
\SetKwInput{KwOutput}{Output}              %
\SetKwInput{KwInitialize}{Initialize}      %

\SetCommentSty{mycommfont}
\SetKwComment{Comment}{$\triangleright$\ }{}

\usepackage{breqn}
\usepackage{booktabs}
\usepackage{makecell}
\usepackage{multirow}
\usepackage{nicematrix}
\usepackage{multirow}
\usepackage{colortbl}
\usepackage{makecell}
\usepackage{array}

\newcolumntype{L}[1]{>{\raggedright\let\newline\\\arraybackslash\hspace{0pt}}m{#1}}
\newcolumntype{C}[1]{>{\centering\let\newline\\\arraybackslash\hspace{0pt}}m{#1}}
\newcolumntype{R}[1]{>{\raggedleft\let\newline\\\arraybackslash\hspace{0pt}}m{#1}}
\usepackage{sidecap, caption}
\definecolor{cvprblue}{rgb}{0.21,0.49,0.74}
\usepackage[pagebackref,breaklinks,colorlinks,citecolor=cvprblue]{hyperref}
\hypersetup{
    colorlinks=true,
    allcolors=cvprblue,
    }

\usepackage[capitalize,nameinlink]{cleveref}
\crefdefaultlabelformat{#2\textbf{#1}#3}
\crefname{equation}{}{}
\creflabelformat{equation}{#2\textup{\bf #1}#3}
\crefname{equation}{Eq.}{Eqs.}
\Crefname{equation}{Equation}{Equations}
\crefname{figure}{Fig.}{Figs.}
\Crefname{figure}{Figure}{Figures}
\crefname{table}{Tab.}{Tabs.}
\Crefname{table}{Table}{Tables}
\crefname{section}{Sec.}{Secs.}
\Crefname{section}{Section}{Sections}
\crefname{problem}{Problem}{Problems}
\Crefname{problem}{Problem}{Problems}
\crefname{definition}{Definition}{Definitions}
\Crefname{definition}{Definition}{Definitions}
\crefname{lemma}{Lemma}{Lemmas}
\Crefname{lemma}{Lemma}{Lemmas}
\crefname{theorem}{Thm.}{Thms.}
\Crefname{theorem}{Theorem}{Theorems}
\crefname{remark}{Rmk.}{Rmks.}
\Crefname{remark}{Remark}{Remarks}
\crefname{enumi}{item}{items}
\Crefname{enumi}{Item}{Items}
\crefname{algocf}{Alg.}{Algs.}
\Crefname{algocf}{Algorithm}{Algorithms}
\crefname{assumption}{Asm.}{Asms.}
\Crefname{assumption}{Assumption}{Assumptions}
\crefname{ALC@unique}{line bla}{lines}
\Crefname{ALC@unique}{Line bla}{Lines}

\newlist{rquestions}{enumerate}{1}
\setlist[rquestions,1]{
    label={\bf RQ\arabic*:},
    ref=\arabic*, %
    labelwidth=!,
    align=left,
    itemindent=0pt,
    leftmargin=30pt,
    }
\creflabelformat{rquestionsi}{#2#1#3} %
\crefname{rquestionsi}{research question number}{research questions number} %
\Crefname{rquestionsi}{Research question number}{Research questions number}

\newcommand{\best}[1]{\textcolor{teal}{\bf #1}}
\newcommand{\second}[1]{\textcolor{orange}{\bf #1}}

\definecolor{cvprblue}{rgb}{0.21,0.49,0.74}
\usepackage[pagebackref,breaklinks,colorlinks,citecolor=cvprblue]{hyperref}

\def\paperID{145} %
\def\confName{3DV\xspace}
\def\confYear{2025\xspace}

\newcommand{\titleTemplate}{%
SurfR:\@ Surface Reconstruction with Multi-scale Attention
}
\newcommand{\authorsTemplate}{%
Siddhant Ranade\textsuperscript{*}
\\ University of Utah \and
Gon\c{c}alo Dias Pais\thanks{Equal contribution.}
\\ Instituto Superior T\'ecnico \and 
Ross Tyler Whitaker
\\ University of Utah \and
Jacinto C. Nascimento 
\\ Instituto Superior T\'ecnico \and
Pedro Miraldo
\\ Mitsubishi Electric\\ Research Laboratories \and 
Srikumar Ramalingam
\\ Google Research
}

\begin{document}
\title{\titleTemplate} 
\author{\authorsTemplate}
\date{\instituteTemplate}
\maketitle
\setcounter{footnote}{0}
\begin{abstract}
We propose a fast and accurate surface reconstruction algorithm for unorganized point clouds using an implicit representation. 
Recent learning methods are either single-object representations with small neural models that allow for high surface details but require per-object training or generalized representations that require larger models and generalize to newer shapes but lack details, and inference is slow.
We propose a new implicit representation for general 3D shapes that is faster than all the baselines at their optimum resolution, with only a marginal loss in performance compared to the state-of-the-art. We achieve the best accuracy-speed trade-off using three key contributions. Many implicit methods extract features from the point cloud to classify whether a query point is inside or outside the object. First, to speed up the reconstruction, we show that this feature extraction does not need to use the query point at an early stage (lazy query). Second, we use a parallel multi-scale grid representation to develop robust features for different noise levels and input resolutions. Finally, we show that attention across scales can provide improved reconstruction results.
\end{abstract}
    
\section{Introduction}\label{sec:introduction}

\begin{figure}[t]
    \centering
    \begin{subfigure}[b]{0.75\linewidth}
    \includegraphics[width=0.49\linewidth]{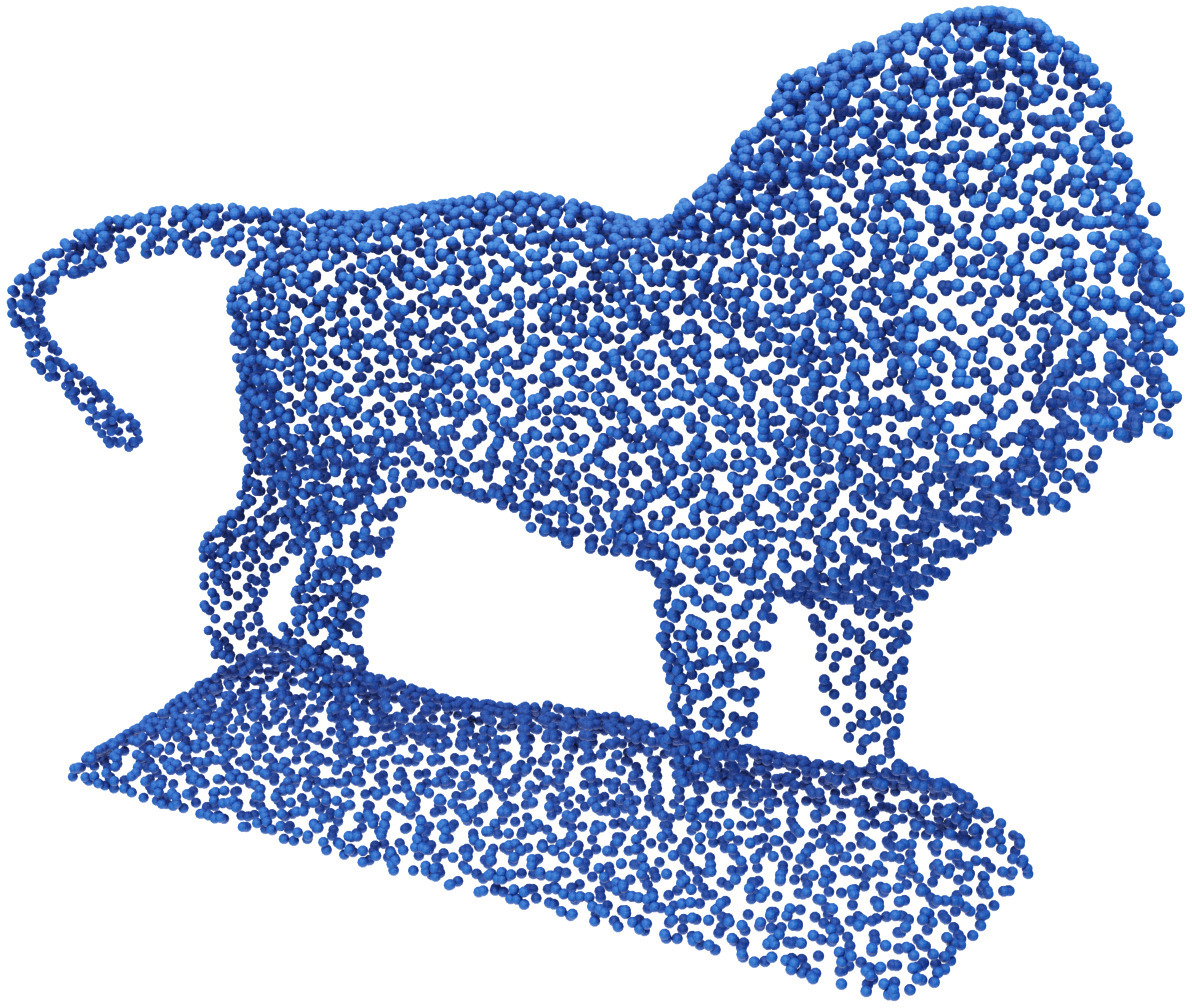}
    \includegraphics[width=0.49\linewidth]{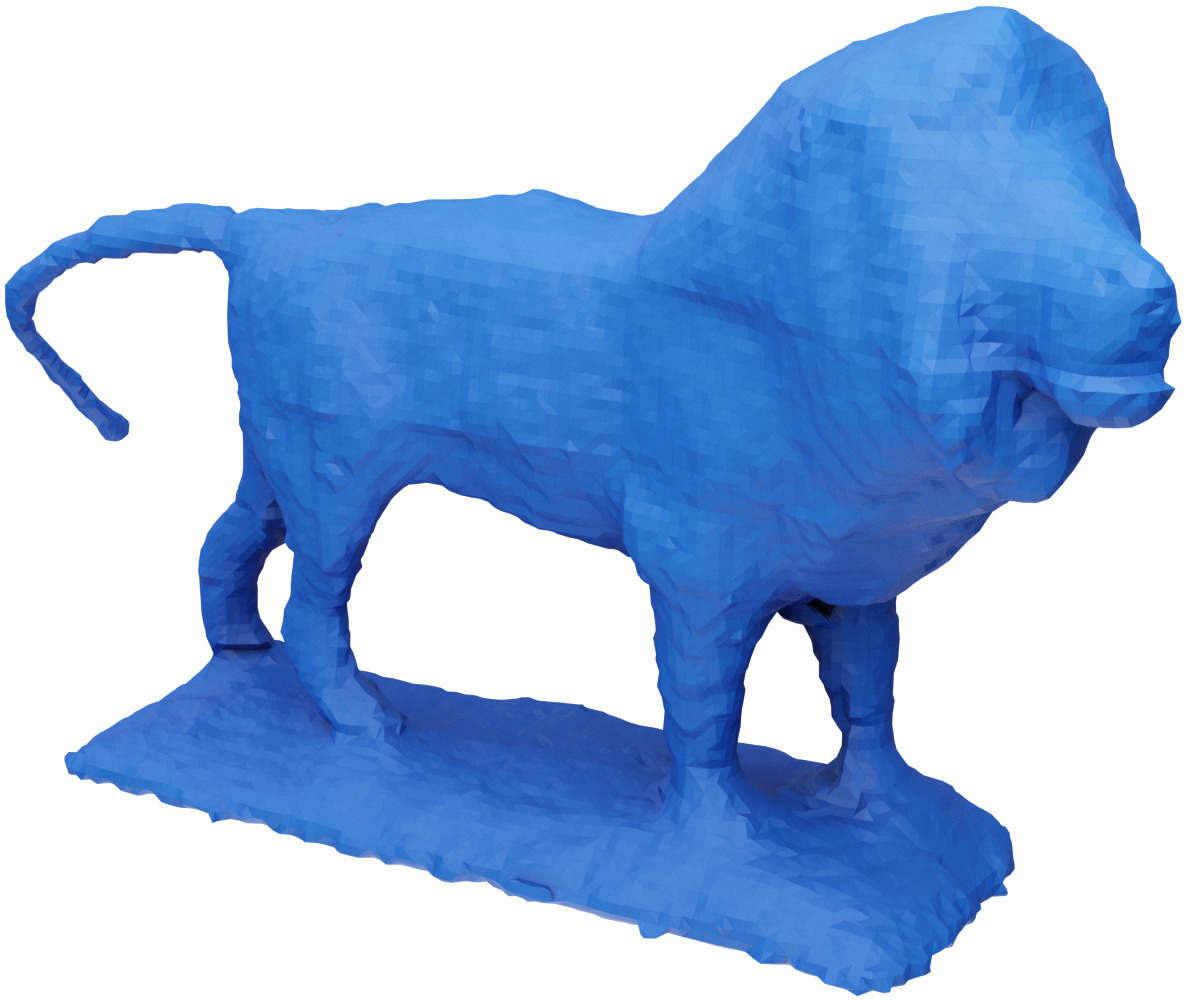} 
    \end{subfigure} \\\vfill
    \begin{subfigure}[b]{.85\linewidth}
    \includegraphics[width=1\linewidth]{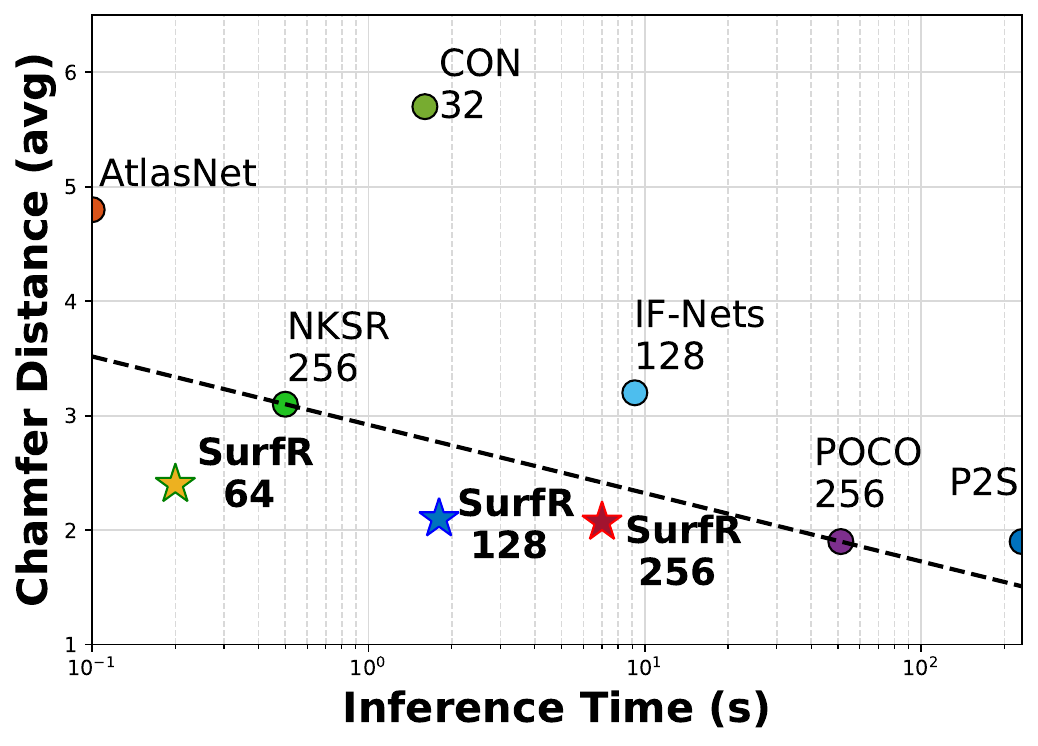}
    \end{subfigure}
    \caption{{\bf SurfR result and error/time trade-off}: At the top, we show a sparse input point cloud and its reconstruction using the proposed method {\bf SurfR}. At the bottom, we show our method and the baselines {\bf error vs. time trade-off} comparison, with different resolutions (number below).}
    \label{fig:teaser}
\end{figure}

Surface reconstruction from noisy point clouds (e.g., from structure-from-motion pipelines or 3D sensors) has been one of the essential tasks in the computer vision and graphics community~\cite{carr_reconstruction_2001,Kolluri2004,Labatut2009RobustAE,kazhdan_screened_2013}, with numerous applications in autonomous vehicles, medical imaging, and AR/VR technologies. The challenges arise from noisy/incomplete data, the choice of representation, complex geometry, and computational bottlenecks. While 3D point-based representations are lightweight, they lack surface information, and voxels, on the other hand, leave a huge memory footprint. We propose {\bf SurfR}, a data-driven and efficient implicit surface reconstruction method that produces high-fidelity models, striking the right balance between speed and accuracy (see \cref{fig:teaser}). Note that applications that require real-time processing or have low compute budgets will benefit immensely from this speedup, including those in robotics, augmented/virtual reality, and embedded systems, for example, in interactive mesh editing, such as \cite{yu_2004,sorkine_2004}, where re-meshing is frequently re-run, or interactive SLAM \cite{Nashed_2018,sidaoui_2019}, where the user accompanied by a robot interacts with the environment to get accurate object representations for scene understanding~\cite{nicholson2018quadricslam,hall2020probabilistic}. SurfR proposes a framework that is towards real-time without compromising surface details and without over-smoothing.

A robust surface reconstruction method should be capable of handling noisy/partial point clouds sampled from a variety of shapes. 
Neural implicit functions show promising results on various 3D tasks, including surface reconstruction. Many of these methods can be divided as per-object or general shape reconstruction. More object-specific methods~\cite{takikawa2021neural,gaur2024oriented}, inspired by~\cite{mildenhall2020nerf} and~\cite{peng2020convolutional}, take advantage of the object's shape priors to encode the objects explicitly. However, these methods are not generalizable to unseen shapes. On the other hand, general shape reconstruction methods~\cite{park_deepsdf_2019,peng2020convolutional,chibane2020implicit,takikawa2021neural,williams2022neural,huang2023neural} rely on general encoders to reconstruct the 3D implicit surface. These methods are more generalizable but suffer from a loss in reconstruction quality.
The general idea is to use a neural network to predict a distance value given a query point based on a feature representation. One can achieve this by first extracting a global feature (e.g., using a PointNet~\cite{pointnet_2017}) and then regressing the distance function for the query point~\cite{park_deepsdf_2019}. However, using a single global feature for the entire point cloud fails to capture the local geometry, and recent methods extract multiple features from local patches~\cite{chibane2020implicit,erler_points2surf_2020,boulch2022poco}. The proposed algorithm also uses multiple resolution features to better model the local and global information to achieve high-fidelity implicit representation. Our contributions are:
\begin{itemize}
\item Lazy query processing in feature extraction to achieve significant computational efficiency.
\item Parallel multi-scale feature extraction to achieve robustness to noise and different input resolutions.
\item Attention using a transformer encoder layer~\cite{vaswani2017transformers} to fuse features across scales, significantly improving reconstruction results. 
\item Best accuracy-speed trade-off. SurfR is competitive with state-of-the-art reconstruction fidelity and is faster than baselines at their optimal resolution.
\end{itemize}

\section{Related Work}

Surface reconstruction spans several decades of research in geometry processing, addressing a challenging and long-standing problem. We can study these methods across different dimensions: generalizable (single model for multiple objects), feature extraction (single point feature or multi-scale features), method architecture (query feature regression or learned kernels), etc.

\paperpar{Generability} 
Learning-based approaches for surface reconstruction take advantage of rich data priors to better handle noise, incomplete data, and inconsistent normals, which can be separated into object-specific and generalizable methods.
Object-specific methods~\cite{Dai_2019_CVPR,Liao2018,wang2018pixel2mesh,mildenhall2020nerf,park_deepsdf_2019,mescheder2019occupancy,pifuSHNMKL19,Meshry2019,sitzmann2019deepvoxels,chabra_deep_2020,tancik2020fourier,groueix_papier-mache_2018,sitzmann2020implicit,Dai_2020_CVPR,Jiang_2020_CVPR,takikawa2021neural,peng2020convolutional,martel2021acorn,gaur2024oriented} focus on a single object/scene to create an implicit surface representation from the input point cloud that is decoded when necessary. These methods can lead to high-quality reconstructions, though they are limited in generalizing newer shapes and geometry. 
In contrast, generalizable methods such as~\cite{park_deepsdf_2019,erler_points2surf_2020,peng2020convolutional,chibane2020implicit,chibane2020ifnet_texture,boulch2022poco,wang2023alto,williams2021neural,williams2022neural,huang2023neural} can generalize to a wide range of shapes, with a penalty in the reconstruction quality. These typically differ in how the query representation is built. The proposed method follows the latter.

\paperpar{Feature Extraction}
In general, all learning methods have a similar structure. A feature representation is built from the input point cloud and then decoded to get its implicit surface. In the object-centric methods, this feature representation is associated with an explicit per-object geometric structure~\cite{peng2020convolutional,takikawa2021neural,gaur2024oriented} known {\it a priori} since the object's geometry is known. The generalizable methods build that representation based on the query object and then decode it to obtain the surface representation. Some methods rely only on a single point feature~\cite{park_deepsdf_2019,erler_points2surf_2020,boulch2022poco,wang2023alto} while others use multiple scale features~\cite{williams2021neural,williams2022neural,huang2023neural}. Multi-scale features have been shown to provide benefits in dealing with noise and different input resolutions for different tasks~\cite{park_deepsdf_2019,chibane2020implicit,chibane2020ifnet_texture,huang2023neural}.
In IF-Nets~\cite{chibane2020implicit,chibane2020ifnet_texture}, a 3D convolutional network (with pooling) processes a discretized point cloud to create feature grids at multiple scales, with the early stages capturing shape detail and the later stages the global structure.
NKSR~\cite{huang2023neural} uses a sparse UNet to obtain per-resolution features.
In contrast, SurfR uses a PointNet for each scale to compute cell features in parallel, considering only points within the cell without discretization. Cell embedding is sensitive to local detail at larger scales, giving us a multi-scale 3D representation. The cross-scale attention balances feature extraction between scales. The contributions are this way of extracting multi-scale features and fusing them with attention. Also, avoiding 3D convolutions and discretization makes SurfR faster and more accurate.

\paperpar{Architecture}
These neural models can be constructed in two ways: estimate the implicit representation from anchors following the estimated kernels or regression from an explicit feature representation.
Kernel methods like~\cite{williams2021neural,williams2022neural,huang2023neural} create a grid kernel from the input point cloud that predicts the implicit representation. The output representation of the input points is the sum of all neighbor grid voxels output weighted by the learned kernel. In~\cite{williams2022neural}, NKS proposes a single-scale kernel to predict the grid occupancy while considering noise. In~\cite{huang2023neural}, NKSR extends NKS for multi-resolution kernels conditioned by UNet features.

On the other hand, the regression methods~\cite{park_deepsdf_2019,erler_points2surf_2020,chibane2020implicit,chibane2020ifnet_texture,boulch2022poco,wang2023alto} create an intermediate feature representation and then decode it to obtain the surface representation.
Using a sparse convolutional backbone, POCO~\cite{boulch2022poco} encodes the input points in feature vectors. Given an arbitrary query point, its feature is computed from the nearest input points' features. POCO proposes to use an attention module to weigh the contribution of each neighbor feature to the query point. ALTO~\cite{wang2023alto} extends POCO's approach and builds successive triplanes~\cite{chan2022efficient,bahat2022neural} to query with attention to the points' location for better feature extraction. In contrast, the proposed method also encodes the input points in feature vectors but uses a PointNet~\cite{pointnet_2017} for each resolution and utilizes an attention layer per resolution. The feature of the query point is constructed from the different resolutions weighted by the inverse distance of the query point to the input points. This scheme allows cross-scale information sharing without over-smoothing and at the cost of details.

\section{Surface Reconstruction with Multi-scale Attention}
Given a set of unorganized points $\mathcal{P}$, the goal of SurfR is to reconstruct the original object surface using implicit surface reconstruction with an SDF. Then, we define the surface $\mathcal{S}$ as the zero level-set of its SDF $d_{\mathcal{S}}(\textbf{x})$, i.e., $\mathcal{S} = \{\textbf{x}, d_{\mathcal{S}}(\textbf{x}) = 0\}$, where $\textbf{x}$ is a point in 3D. We are essentially learning the implicit function using a neural network in order to produce SDF estimates. Similar to other implicit neural methods, we estimate the SDF on a regular grid of desired resolution, and a suitable isosurfacing algorithm is necessary to recover the underlying surface~\cite{lorensen1987marching}. The overview of the method is shown in~\cref{fig:arch}.

\subsection{Overview and Notation}
\label{sec:overview}

\newcommand{\figscale}{0.95}

\begin{figure}[t]
    \centering
    \resizebox{\linewidth}{!}{
        \includegraphics{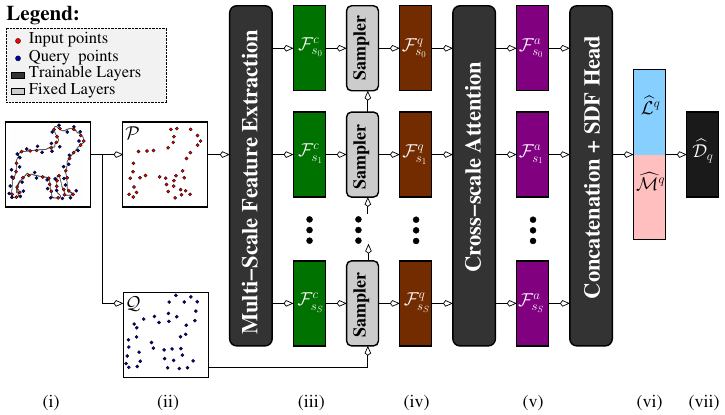}
    }
    \caption{\textbf{Overview} The input consists of point cloud $\mathcal{P}$ and query points $\mathcal{Q}$. For each scale $s \in \{s_0, s_1, \dots, s_S\}$, an encoder network is used to extract per-cell features $\mathcal{F}^c_s$. These features are sampled at the query points to get query point features $\mathcal{F}^q_s$  with our novel query-feature sampling technique. A self-attention mechanism is applied to these features to obtain a new set of per-query features $\mathcal{F}^a_s$, which are subsequently concatenated and regressed by the SDF head to sign logits $\hat{\mathcal{L}}^q$ and magnitudes $\hat{\mathcal{M}}^q$, which are used to compute the final signed distance $\hat{\mathcal{D}}_q$.\vspace{0pt}}
    \label{fig:arch}
\end{figure}

\paperpar{Overview}
We highlight the five main components of SurfR in this section.
First, we extract features at multiple scales from the input point cloud. The basic idea is to partition the space into cells and then extract features from the points lying in each of these cells. Second, we lazily bring the query points and use the proposed query feature sampler to extract query features. Third, a cross-scale attention is applied to these features using a transformer layer. Next, we concatenate the features across scales and use the SDF head to regress the signed distance. Finally, we use the loss functions on the sign and magnitude, and an L1 penalty on weights for regularization. In the subsections that follow, we discuss the individual components of the algorithm in detail.

\paperpar{Input and groundtruth}
The proposed method takes as input the point cloud $\mathcal{P}=\{\textbf{p}_i: i \in \{0, \ldots, P-1\}\}$ and the query points $\mathcal{Q}=\{\textbf{q}_j: j \in \{0, \ldots, Q-1\}\}$, where $\textbf{p}_i,\textbf{q}_j\in\mathbb{R}^3$, shown in~\cref{fig:arch} (i) and (ii). For training, we use the ground-truth signed distances (i.e.\ to the mesh) $\mathcal{D}^q=\{d^q_j : j \in \{0, \ldots, Q-1\}\}$ corresponding to each query point. All input and query points are normalized to lie inside the unit cube $[-1, 1]^3$.

\paperpar{Features}
We extract features at different stages of the algorithm, and we briefly introduce the notation here. For each scale $s \in \{s_0, s_1, \dots, s_{S}\}$, we extract per-cell features $\mathcal{F}^c_s$, where the superscript $c$ denotes the cell and the subscript $s$ denotes the scale. The query features are denoted by $\mathcal{F}^q_s$. The features generated through cross-scale attention are denoted by $\mathcal{F}^a_s$. The scale $s$ is usually used as the subscript in different entities that follow.

\paperpar{Encoders} The local and cell encoders at scale $s$ are denoted by $e^l_s$ and $e^c_s$, and the attention across scales by $e^a$.

\subsection{Parallel Multi-scale Feature Extraction} \label{ss:msfe}

\begin{figure}[t]
        \centering
        \resizebox{\figscale\linewidth}{!}{\includegraphics{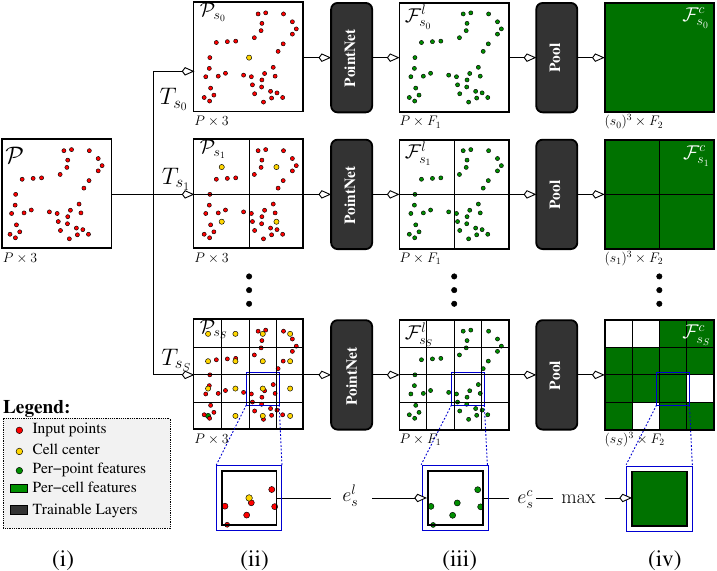}}
        \caption{\textbf{Parallel Multi-scale feature extraction} The input points $\mathcal{P}$ in (i) are transformed to cell coordinates $\mathcal{P}_s$, at each scale $s$ (ii), and each individual cell is processed in parallel as shown in the inset. Then, the points are encoded by PointNet $e^l_s$ to get the per-point features $\mathcal{F}^l_s$ (iii). Further they are encoded by the cell encoder $e^c_s$ and pooled cell-wise to get the per-cell features $\mathcal{F}^c_s$ (iv).
        }
        \label{fig:msfe}
\end{figure}
\begin{figure}[t]
        \centering
        \resizebox{\figscale\linewidth}{!}{\includegraphics{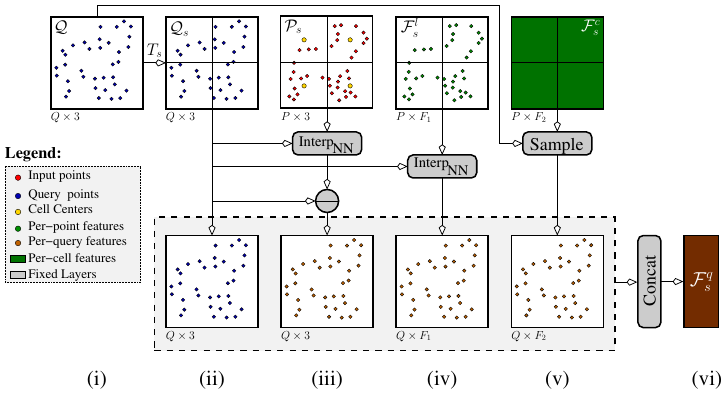}}
        \caption{\textbf{Query Feature Sampling} The input to this stage consists of $\mathcal{P}_s$, $\mathcal{F}^l_s$, $\mathcal{F}^c_s$ from the multi-scale feature extraction, and the query points $\mathcal{Q}$. The query points \emph{ii} are transformed to cell coordinates $\mathcal{Q}_s$ (\emph{ii} - top). The output per-query feature $\mathcal{F}^q_s$ (\emph{vi}) is a concatenation of the query points in cell coordinates  (\emph{ii} - bottom), mean relative nearest neighbor position (\emph{iii}),  mean nearest neighbor feature (\emph{iv}), and feature of the cell containing the query point (\emph{v}).}
        \label{fig:qfs}
\end{figure}

\Cref{fig:msfe} presents the proposed approach to multi-scale feature extraction from a given point cloud. At each scale, the input point set is spatially partitioned into grid cells, and points within each cell are processed independently of other cells, as shown in the inset. At a larger scale, we partition the space into a larger number of cells and vice versa. Before extracting features from the points falling in each of these cells, we transform the points to the local coordinate frame of the cells. As shown in \cref{fig:msfe}~(ii), the points $\mathbf{x}$ at scale $s$ are first transformed to cell coordinates using $T^s(\mathbf{x})$, which corresponds to a translation of the origin to the cell center and scaling. Then, a per-point feature (\cref{fig:msfe}~(iii)) is extracted using a PointNet $e^l_s$. Finally, a single feature is computed using the cell encoder $e^c_s$ followed by cell-wise max pooling. The result is shown in \cref{fig:msfe}~(iv). Next, we detail these steps.

\paperpar{Transformation to cell coordinates $T_s(\mathbf{x})$} Transforming the points to cell coordinates allows the local encoder to independently estimate features across cells. At scale $s$, the unit cube is divided into a total of $s^3$ cells. $T_s(\mathbf{x})$ denotes the cell-coordinate of point $\mathbf{x}$ at scale $s$.

\paperpar{Per-point features}
Let $\mathcal{P}_s = \{T_s(\textbf{p}_i): i \in \{0, \ldots, P-1\}\}$ denote the set of points in local coordinates at scale $s$. For each point $\mathbf{p}_i \in \mathcal{P}_s$, the local encoder $e^l_s$ is used to extract per-point features as follows:
\begin{equation}
    \label{eq:per_point_features}
    \mathbf{f}^{l,i}_s = e^l_s(T_s(\mathbf{p}_i)),
\end{equation}
which is used by the query feature sampler in computing the mean nearest neighbor feature in \cref{ss:qfs}. We denote the set of all per-point features at scale $s$ as $\mathcal{F}^l_s = \{\mathbf{f}_s^{l,i}: \mathbf{p}_i \in \mathcal{P}\}$.

\paperpar{Per-cell features} The per-point features $\mathbf{f}_s^{l,i}$ are further encoded by the cell encoder $e^c_s$ and pooled over cells to get the per-cell features using 
\begin{equation}
    \mathbf{f}^c_s= \max_{i \in c_k} e^c_s(\mathbf{f}^{l,i}_s),
    \label{eq:per_cell_features}
\end{equation}
where $\max$ leads to a channel-wise maximum overall point in cell $c_k$. These per-cell features account for the local geometry of the cell, which is crucial for estimating the SDF. Cells that do not contain any points have the all-zero feature, shown in white in \cref{fig:msfe}~(iv). The set of all per-cell features (also at scale $s$) are denoted as $\mathcal{F}^c_s = \{\mathbf{f}^c_s: c \in \{0, \ldots, s-1\}^3\}$.

\subsection{Query Feature Sampling}
\label{ss:qfs}

Both local and global shape information are critical in estimating the SDF at a given point. Methods that use a single global feature descriptor, e.g.~\cite{park_deepsdf_2019,groueix_papier-mache_2018}, do not perform as well as methods that take the local shape into account. One way to incorporate the local shape is to use patches around query points as input. However, this is expensive. The proposed architecture addresses this issue by first extracting shape features independently of the query points and then \textit{samples} these features at the query points. At each scale, the set of per-point features $\mathcal{F}^l_s$, (as shown in \cref{eq:per_point_features}), and per-cell features $\mathcal{F}^c_s$ (see  \cref{eq:per_cell_features}) are combined to obtain features $\mathcal{F}^q_s$ for each query point, as shown in \cref{fig:qfs} (\cpageref{fig:qfs}). The feature $\mathbf{f}^{q,j}_s$ corresponding to a query point $\mathbf{q}_j$ lying in cell $s$ is defined as the concatenation of:
\begin{itemize}
    \item {\bf Query point position}: $T_s(\mathbf{q}_j)$.
    \item {\bf Cell feature}: $\mathbf{f}^c_s$.
    \item {\bf Mean nearest neighbor feature} Given ${\cal N}_j \subseteq \{i:\mathbf{p}_i \in {\cal P}^c_s\}$, the set of $K$ nearest neighbors of $\mathbf{q}_j$, the mean nearest neighbor feature is given by:
    \begin{equation}
        \mathbf{f}^{{\cal N}_j}_s = \sum_{i \in {\cal N}_j}w_s^{j,i} \mathbf{f}^{l,i}_s,
        \label{eq:mean_nearest_neighbor}
    \end{equation}
     where the weights $w_s^{j,i} = \frac{\text{sim}(i,j)}{\sum_{l \in {\cal N}_j} \text{sim}(l,j)}$, $\sum_i w_s^{j,i} = 1$ ({\bf InterpNN}), and $\text{sim}(i,j)=\frac{1}{\|\mathbf{q}_j - \mathbf{p}_i\|^2}$, similar to~\cite{qi_pointnet_2017}.
    \item {\bf Mean relative nearest neighbor position} a weighted average of the relative positions of the $K$ nearest neighbors in cell coordinates, i.e.,
    \begin{equation}
    \bar{\mathbf{p}}^{{\cal N}_j}_s = 
    s\sum_{i \in {\cal N}_j} w_s^{j,i} (\mathbf{p}_i - \mathbf{q}_j).
    \label{eq:mean_relative_nn}
    \end{equation}
\end{itemize}

The final concatenated feature of $\mathbf{q}_j$ in cell $c$ is given by
\begin{equation}
\label{eq:query_feature}
\mathbf{f}^{q,j}_s = \left[ T_s(\mathbf{q}_j),\  \mathbf{f}^c_s,\ \mathbf{f}^{{\cal N}_j}_s,\  \bar{\mathbf{p}}^{{\cal N}_j}_s \right].
\end{equation}
The set of all per-query features at scale $s$ is denoted as $\mathcal{F}^q_s = \{\mathbf{f}^{q,j}_s: j \in \{0, \ldots, Q-1\}\}$.

\subsection{Cross-Scale Attention} \label{ss:att}

The per-query features from different scales contain complementary information. Therefore, we adopt a weighted sum across all scales to emphasize the more significant features from different scales and improve feature extraction. For that purpose, a cross-scale attention mechanism is included after the query feature sampling. For each query point, we apply a transformer~\cite{vaswani2017transformers} encoder layer to fuse the features across levels, here denoted as $e^a$. Given the query features $\mathbf{f}^{q,j}_s, \ldots, \mathbf{f}^{q,j}_{s_S}$ as inputs (shown in \cref{eq:query_feature}), the attention mechanism outputs a new set of features $\mathcal{F}^a_s = \{\mathbf{f}^{a,j}_{s_0}, \ldots, \mathbf{f}^{a,j}_{s_S}\}$ as follows:
\begin{equation}
    \mathbf{f}^a_{s,j} =
    e^a(\mathbf{f}^{q,j}_{s}, \{\mathbf{f}^{q,j}_{s_0}, \ldots, \mathbf{f}^{q,j}_{s_S}\}).
    \label{eq:csa}
\end{equation}
These features are shown in \cref{fig:arch}~(v).

As will be shown in \cref{ss:ablation}, we find that this simple self-attention mechanism across scales improves the performance of the method. In particular, it helps reduce the artifacts of the cell boundaries and produces smoother results overall.

\subsection{Signed Distance Regression}
\label{ss:sdr}

The set of features $\mathcal{F}^a_s$, in \cref{eq:csa}, corresponding to query points $\mathbf{q}_j$ are concatenated across scales and regressed by the SDF head $e^h$ to sign logits and magnitudes of the signed distance, denoted by $\widehat{l}^q_j$ and $\widehat{m}^q_j$, respectively (see \cref{fig:arch}(vi)), as
\[
    (\widehat{l}^q_j, \widehat{m}^q_j) = e^h([\mathbf{f}^{a,j}_{s_0}, \ldots, \mathbf{f}^{a,j}_{s_S}]).
\]
The final estimate of the SDF is computed as $\widehat{d}^q_j = \text{sgn}(\widehat{l}^q_j) \cdot \widehat{m}^q_j$, shown in \cref{fig:arch}(vii).

\subsection{Loss Functions} \label{ss:loss}
The loss function minimizes the error on the magnitude and sign of the SDF and uses L1 penalty on the weights for regularisation (similar to~\cite{erler_points2surf_2020}):
\begin{equation}
L = \lambda_{\text{mag}} L_{\text{mag}} + \lambda_{\text{sgn}} L_{\text{sgn}} + \lambda_{\text{reg}} L_{\text{reg}}
\end{equation}
The term $L_{\text{mag}}$ denotes the error on the estimated magnitude with respect to the ground-truth as given by $L_{\text{mag}} = \sum_j \left|\tanh(\hat{m}^q_j) - \tanh(\text{abs}(d^q_j))\right|$. 
The sign term $L_{\text{sgn}}$ is the binary cross-entropy on the sign logits. The term $L_{\text{reg}}$ is an $L_1$ regularizer on the weights of $e^h$. The coefficients $\lambda_{mag}$, $\lambda_{sgn}$, and $\lambda_{reg}$ denote the tunable hyperparameters. 

\subsection{Surface Reconstruction}
\label{ss:reconstruction}
Given a new point cloud as input, we estimate the SDF at points on a regular grid and then use marching cubes to extract the (iso-surface). Estimating the SDF at all the points on a regular grid is prohibitively expensive for any reasonable voxel grid resolution, and instead, one can argue that a truncated SDF is sufficient for surface extraction. As in \cite{erler_points2surf_2020}, we evaluate the SDF close to the input points (leaving other samples empty), and propagate the sign by repeatedly applying a box filter of size $\epsilon^3$ voxels at the empty voxels until convergence, updating the sign only if the filter response is greater (in magnitude) than a user-defined threshold $t_{\text{update}} = 13$. Since $\epsilon$ directly affects reconstruction time, we use the same value, $\epsilon=5$. 

\section{Experiments}

\subsection{Datasets} \label{ss:data}

The training and validation data consist of 4950 and 100 examples from the ABC dataset~\cite{Koch_2019_CVPR}, a collection of CAD models. The test set comprises 100 additional examples from the ABC dataset (unseen in training), the FAMOUS dataset (a collection of 22 meshes popular in the geometry processing literature, such as the Utah teapot, the Stanford bunny, and the Dragon), and the Thingi10K dataset~\cite{Thingi10K}, with varying levels of noise and sparsity (for Thingi10K and FAMOUS). For a fair comparison, we highlight that the test data was not seen during training or validation for any of the tested methods in \cref{ss:results}. We also show qualitative results on two scans of complex real-world objects.

All meshes are first transformed to the unit cube, following which the input point cloud is sampled using BlenSor~\cite{BlenSor}, which simulates scanning of a given surface mesh with a time-of-flight sensor, including realistic noise characteristics and sensor artifacts. For each shape, 2000 query points are used, 1000 samples on the surface and offset along the surface normal by a distance uniformly sampled from [-0.02, 0.02], and 1000 sampled uniformly at random in the unit cube. While training, 6000 input points and 1000 query points were used per shape. During training, random rotations are applied to all shapes.

\subsection{Implementation Details}

\paperpar{Network Architectures} The local encoder $e^l_s$ is implemented as a 2-layer MLP followed by a spatial transformer~\cite{jaderberg_spatial_2015}. We pick the per-point feature size to be $F_1 = 64$. The cell encoder $e^c_s$ is a 3-layer MLP, with the per-cell feature size $F_2=128$. The query feature dimension (at each scale) is thus $F = 3 + 3 + F_1 + F_2 = 198$ (see \cref{fig:qfs}). We use scales $s \in \{1, 4, 16\}$, $S=3$. Together, $e^l_s$ and $e^c_s$ (\cref{eq:per_point_features,eq:per_cell_features}) resemble a PointNet~\cite{pointnet_2017}. The cross-scale attention $e^a$ (\cref{eq:csa}) is composed of one head transformer encoder layer with a 512 fully connected layer. The SDF head $e^h$ has an input dimension of $S \times F = 594$ and is a 5-layer MLP with an $\text{abs}(\cdot)$ non-linearity for the magnitude. All MLPs use BatchNorm~\cite{BatchNorm} followed by ReLU non-linearity for hidden layers.

\paperpar{Training Details} The loss weights are $\lambda_{\text{mag}} = 5.0$, $\lambda_{\text{sgn}} = 2.0$, $\lambda_{\text{reg}} = 10^{-6}$, and the batch size is 16. We use the Adam~\cite{AdamOpt} optimizer with an initial learning rate of $7.5*10^{-4}$, which decays exponentially after every epoch, halving every 100 epochs, and train for a total of 750 epochs.

\paperpar{Software/Hardware} Our code is written in PyTorch~\cite{PyTorch} with the PyTorch-Geometric~\cite{PyTorchGeometric} extension library. We take utmost care to ensure our experiments are reproducible; 
We use one NVidia Titan Xp GPU for training, taking about 20 minutes per epoch.

\subsection{Ablation Study}\label{ss:ablation}

\begin{SCtable}[2][t]
    \caption{{\bf Ablations}: Chamfer distance for: \textbf{Base}: single LOD of value $1$, \textbf{Base + multi-scale}: multi-scales, with values $\{1,4,16\}$, without attention, \textbf{Base + multi-scale + attn.}: multi-scales $\{1,4,16\}$ with cross-scale attention.}
    \label{tab:ablation}
    \centering
    \renewcommand{\arraystretch}{1}
    \resizebox{0.5\linewidth}{!}{%
    \setlength{\tabcolsep}{10pt}\begin{NiceTabular}{>{\kern-\tabcolsep}l<{\kern-\tabcolsep} cc>{\kern-\tabcolsep}c<{\kern-\tabcolsep}}[code-before =%
    \rectanglecolor{Gray!20}{4-1}{4-4}%
    \rectanglecolor{Gray!50}{6-1}{6-4}%
    ]
    \toprule
    & \multicolumn{3}{c}{\thead{Chamfer Distance $L_2$~($\downarrow$)}} \\ \cmidrule(lr){2-4}
    \thead{Noise} & \thead{Base} & \thead{+ multi-scale} & \thead{+ multi-scale \\ + attention}  \\ \midrule
    no & 2.6 & 2.3 & \bf 2.2  \\
    med & 3.0 & 2.7 & \bf 2.4 \\
    max & 3.6 & 3.7 & \bf 3.2 \\
    Avg. & 3.1 & 2.8 & \bf 2.6 \\ 
    \bottomrule
    \end{NiceTabular}
    }
\end{SCtable}

\begin{table}[t]
    \caption{{\bf Comparisons on the Different Multi-scales and Weighting Schemes}: Chamfer distance on different scales with cross-scale attention (lower is better). We also evaluate the different weighting schemes for grid cell aggregation. \textbf{EW} samples equal weights and \textbf{LW} is a learnt weight parameter.}
    \label{tab:ablation_supp}
    \centering
    \setlength{\tabcolsep}{3pt}
    \renewcommand{\arraystretch}{1.2}
    \resizebox{0.9\linewidth}{!}{\begin{tabular}{>{\kern-\tabcolsep}l<{\kern-\tabcolsep} cccc c>{\kern-\tabcolsep}c<{\kern-\tabcolsep}}
    \toprule
    \multirow{2}{*}{\thead{Noise}} & \multicolumn{4}{c}{\thead{InterpNN}} & \thead{LW} & \thead{EW} \\ \cmidrule(lr){2-5}
    &  \{1, 4\}  & \{1, 4, 16\} & \{1, 5, 25\} & \{1, 3, 9, 27\} & \{1, 4, 16\} & \{1, 4, 16\} \\ \midrule
    no & 3.3 & \bf 2.2 & 2.6 & 3.3 & 2.5 & 2.5 \\
    med & 2.8 & \bf 2.4 & 2.7 & 3.2 & 2.6 & 2.8 \\
    max & 3.6 & \bf 3.2 & 3.5 & 3.9 & 3.4 & 3.6 \\
    \rowcolor{Gray!50} Avg. & 3.3 & \bf 2.6 & 2.9 & 3.5 & 2.8 & 3.0 \\
    \toprule
    \end{tabular}}

\end{table}

Using a surface reconstruction of voxel grid of dimension $256$, we evaluate our design choices related to multi-scale feature extraction compared to using a single scale of 1, and query feature components by comparing their performance. We report the chamfer distance on the ABC dataset in \cref{tab:ablation}. This analysis validates the design of the query feature. Due to resource constraints, these ablations were performed with a batch size of $4$, training for 100 epochs. 

In addition to the choices related to multi-scale grid feature extraction and cross-scale attention, we study the effect of using different scales and weighting schemes for query feature sampling. \Cref{tab:ablation_supp} shows the results. We tried with equal {\bf EW} and learned weights {\bf LW} and InterpNN (see \cref{ss:qfs}) which achieves the best results. The final method uses scales \{$1$, $4$, $16$\}, which obtained the best average performance in all ablations. The multi-scale features allow for a better global understanding of the object, while the cross-scale attention parallel module fuses the global and local features, producing more detail. 

\subsection{Results}
\label{ss:results}

\paperpar{Baselines}
We compare the proposed approach with the data-driven methods P2S~\cite{erler_points2surf_2020},  IF-Net~\cite{chibane2020implicit}, CON~\cite{peng2020convolutional}, SAP~\cite{Peng2021}, POCO~\cite{boulch2022poco}, and NKSR~\cite{huang2023neural}. We follow the same training protocol as POCO and P2S. We note that previous methods obtained the best results at different grid resolutions. While CON uses a voxel grid of size 32, IF-Net uses 128, P2S 256, SAP 256, POCO 256, and NKSR 256. CON and IF-Net are trained on $\approx40000$ models across 13 categories from ShapeNet~\cite{shapenet2015}, while P2S, POCO, NKSR, and SurfR are trained on $\approx 5000$ shapes from the ABC dataset. All the baselines except NKSR only require the input point cloud to obtain the surface representation. 
Since NKSR also requires normals as input, we use the evaluation point cloud to estimate the normals with 16 nearest neighbors as a fair comparison to the other baselines. 
Our evaluation scheme does not use additional refinement steps such as POCO and NKSR as detailed in~\cref{ss:reconstruction}.

\begin{figure*}[t]
    \centering
    \setlength{\tabcolsep}{15pt}
    \renewcommand{\arraystretch}{1.3}
    \resizebox{0.85\linewidth}{!}{\begin{tabular}{>{\kern-\tabcolsep}c<{\kern-\tabcolsep} c c c c >{\kern-\tabcolsep}c<{\kern-\tabcolsep}}
    \multirow{2}{*}{\rotatebox[origin=c]{90}{\makecell{\textbf{ABC}\\no-noise}}}
    & \includegraphics[height=0.1\textheight,keepaspectratio]{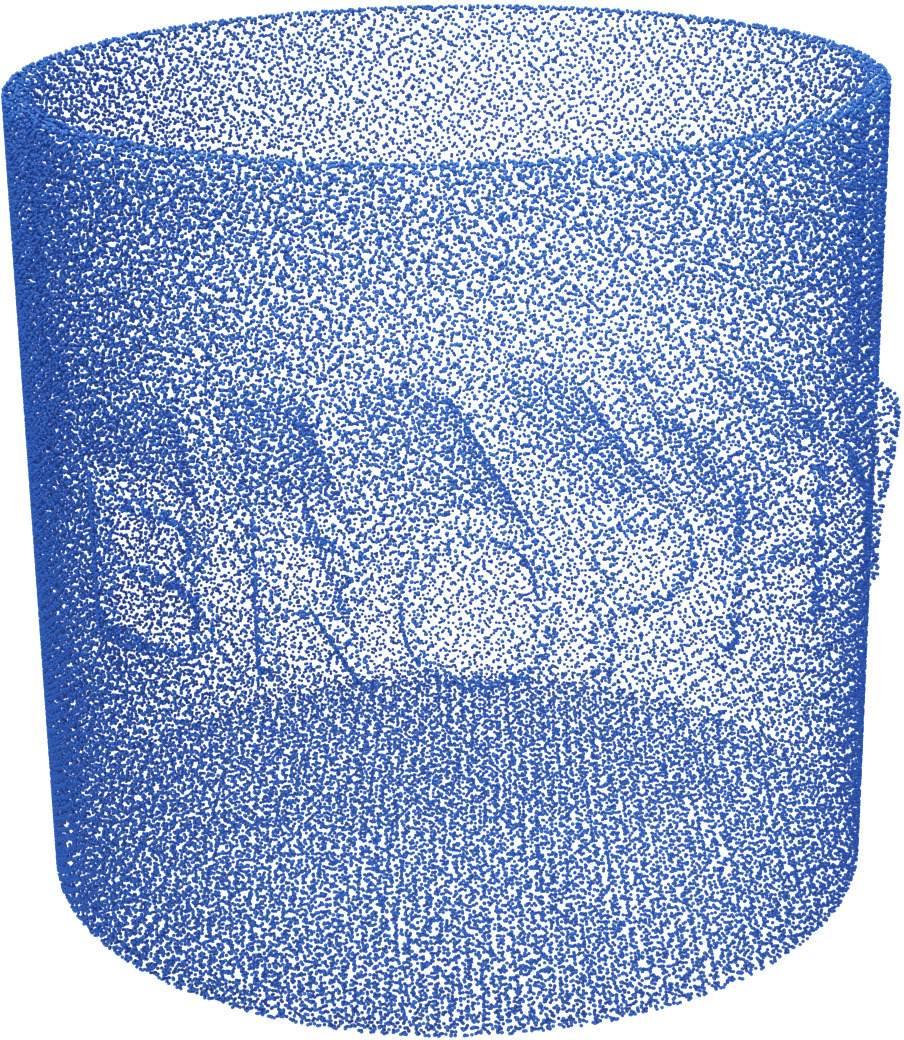}
    & \includegraphics[height=0.1\textheight,keepaspectratio]{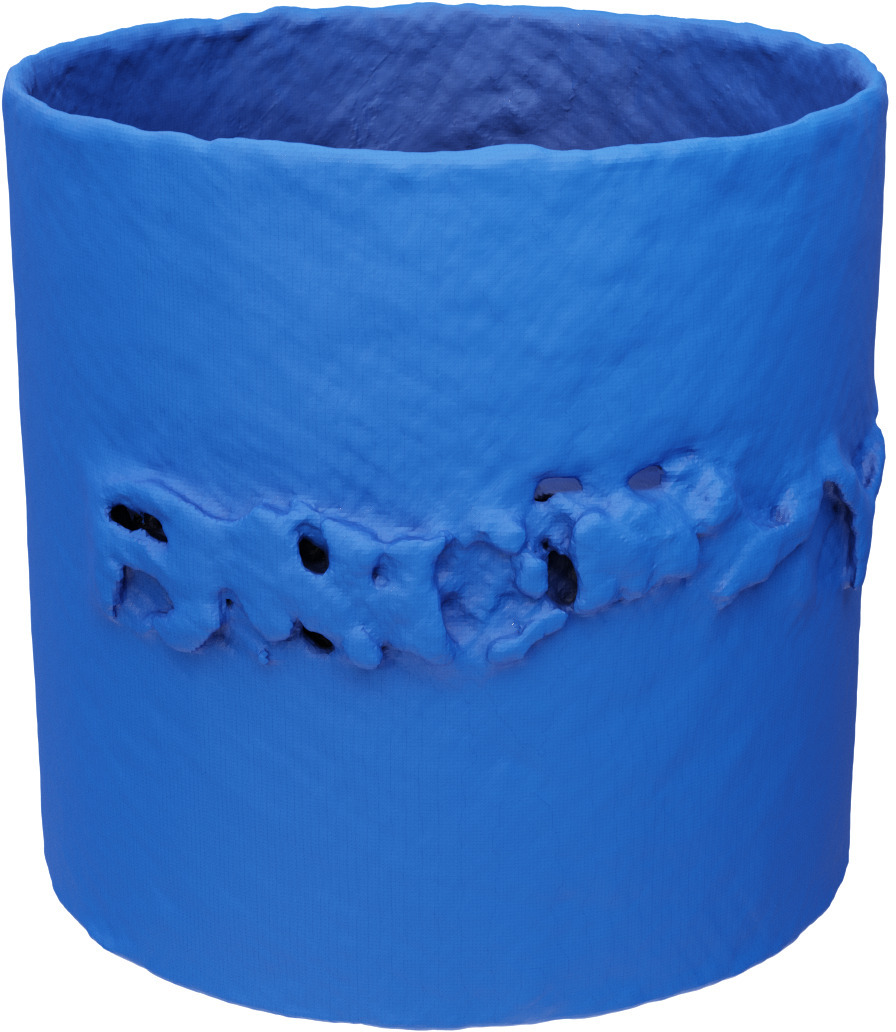}
    & \includegraphics[height=0.1\textheight,keepaspectratio]{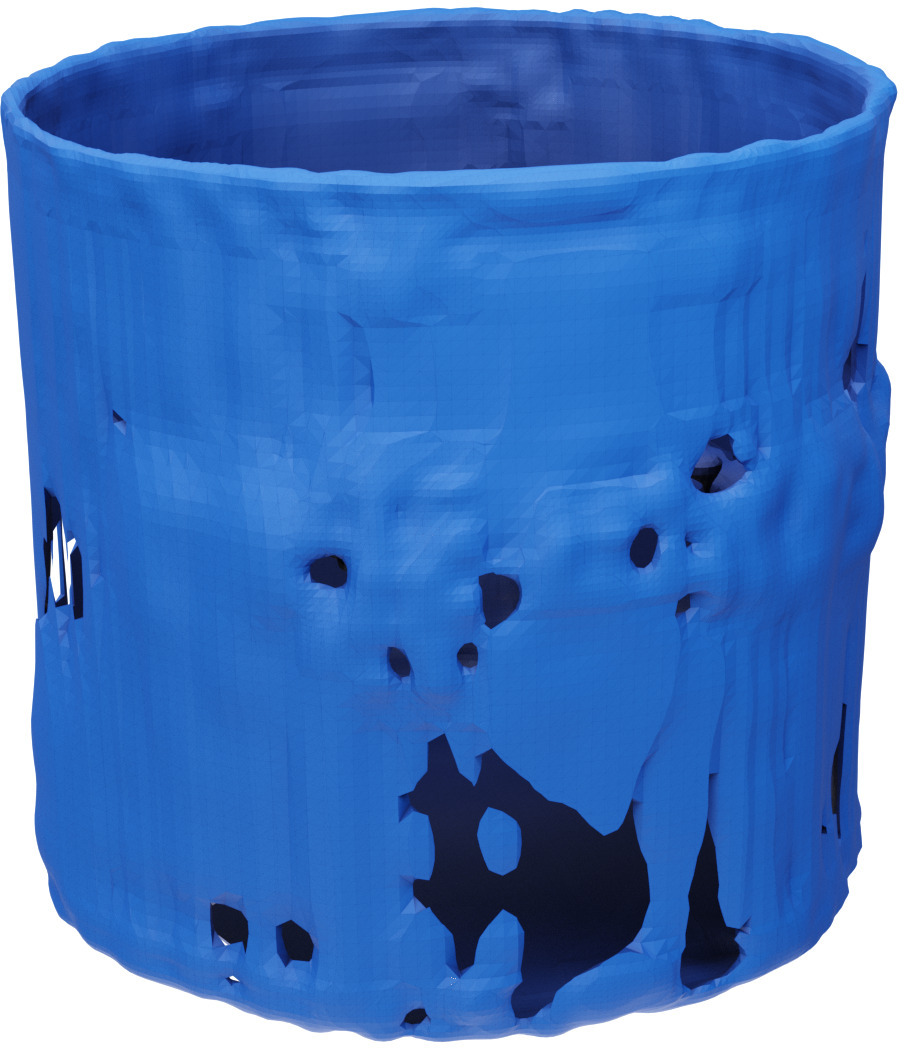}
    & \includegraphics[height=0.1\textheight,keepaspectratio]{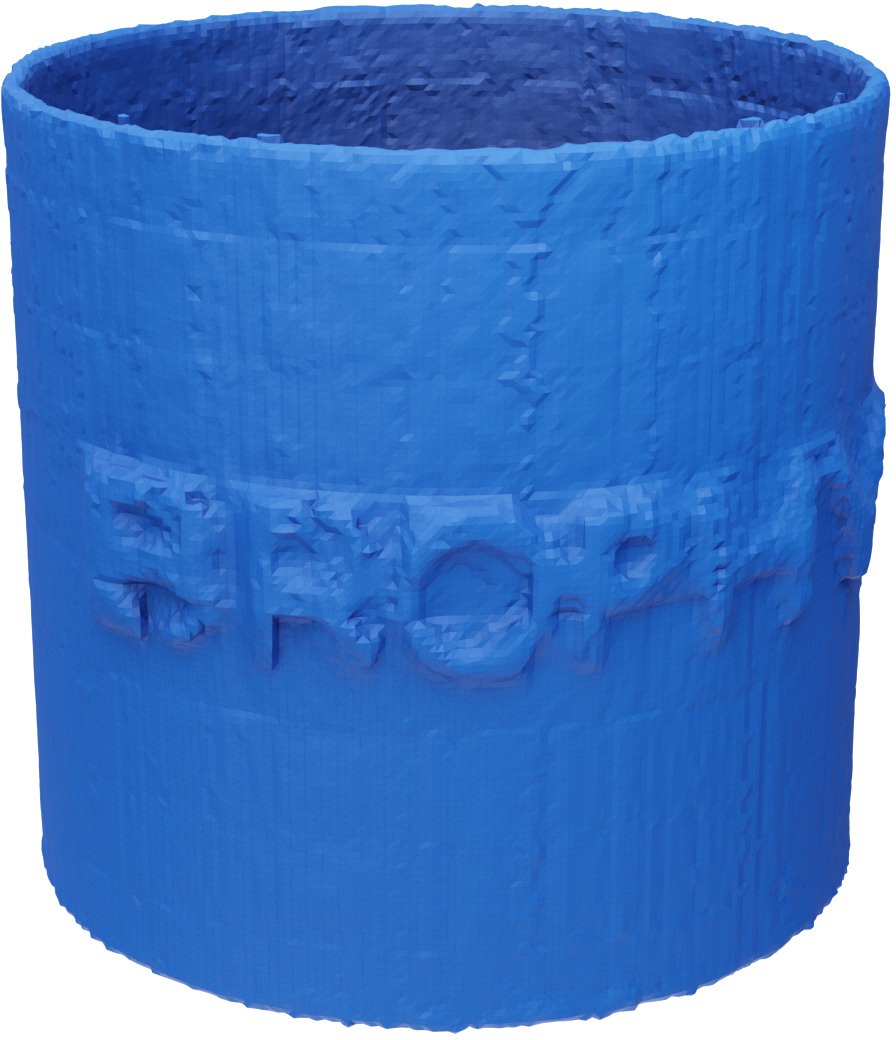}
    & \includegraphics[height=0.1\textheight,keepaspectratio]{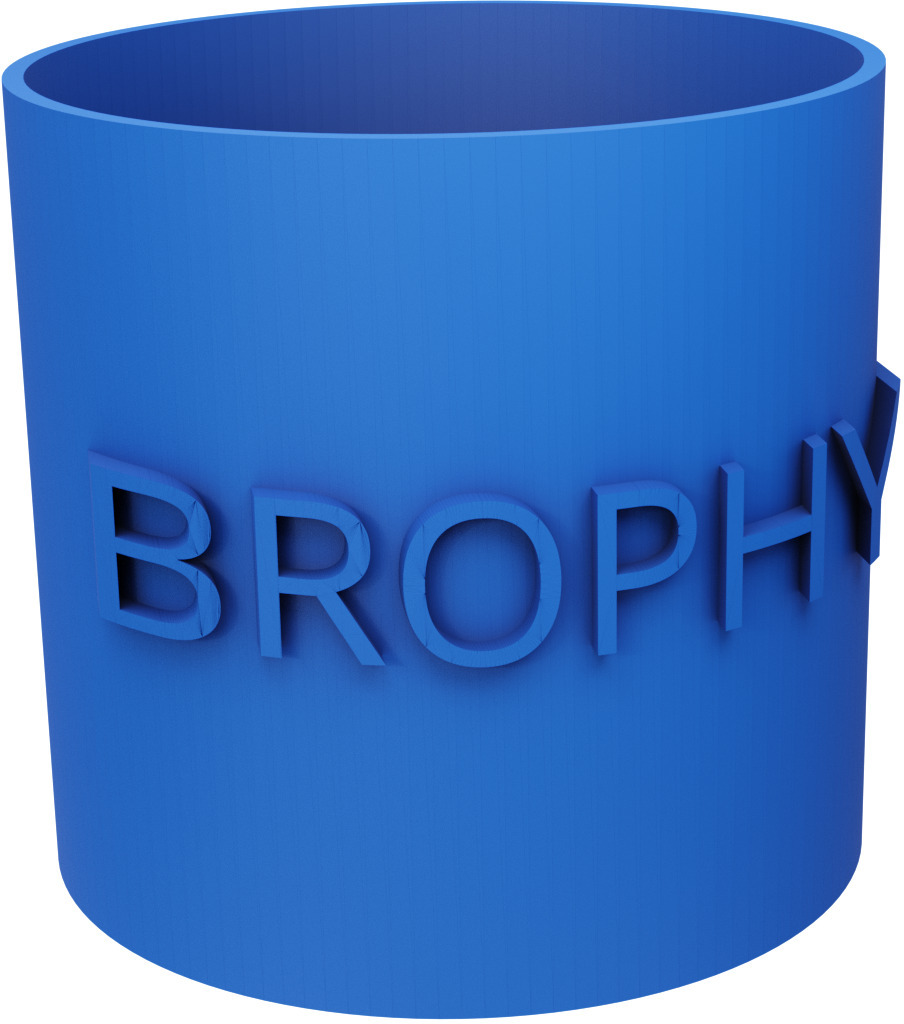} %
    \\
    
    & \includegraphics[height=0.1\textheight,keepaspectratio]{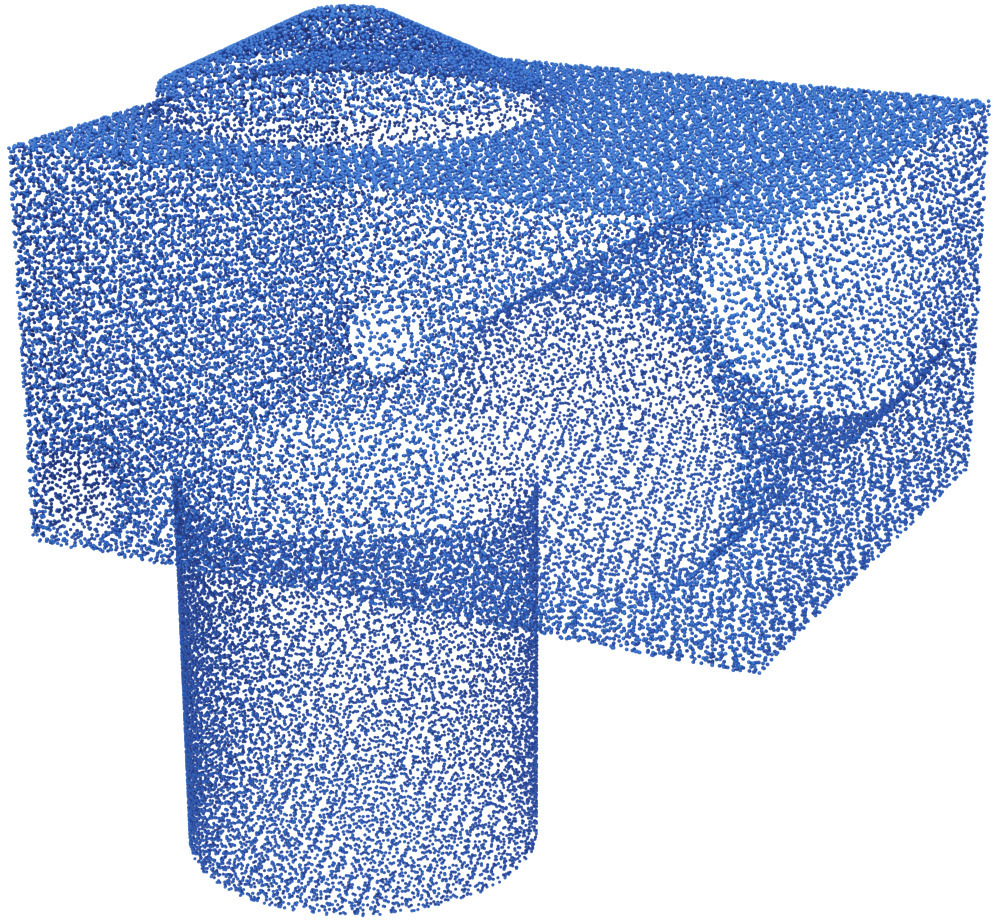}
    & \includegraphics[height=0.1\textheight,keepaspectratio]{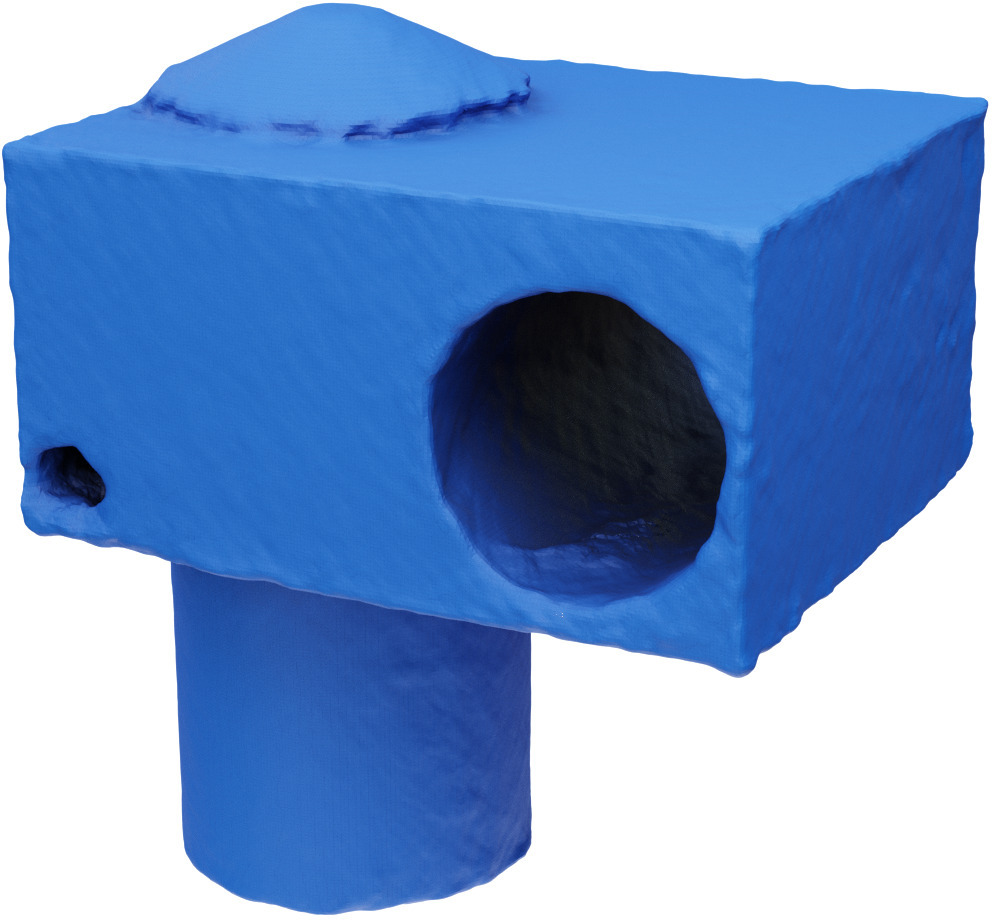}
    & \includegraphics[height=0.1\textheight,keepaspectratio]{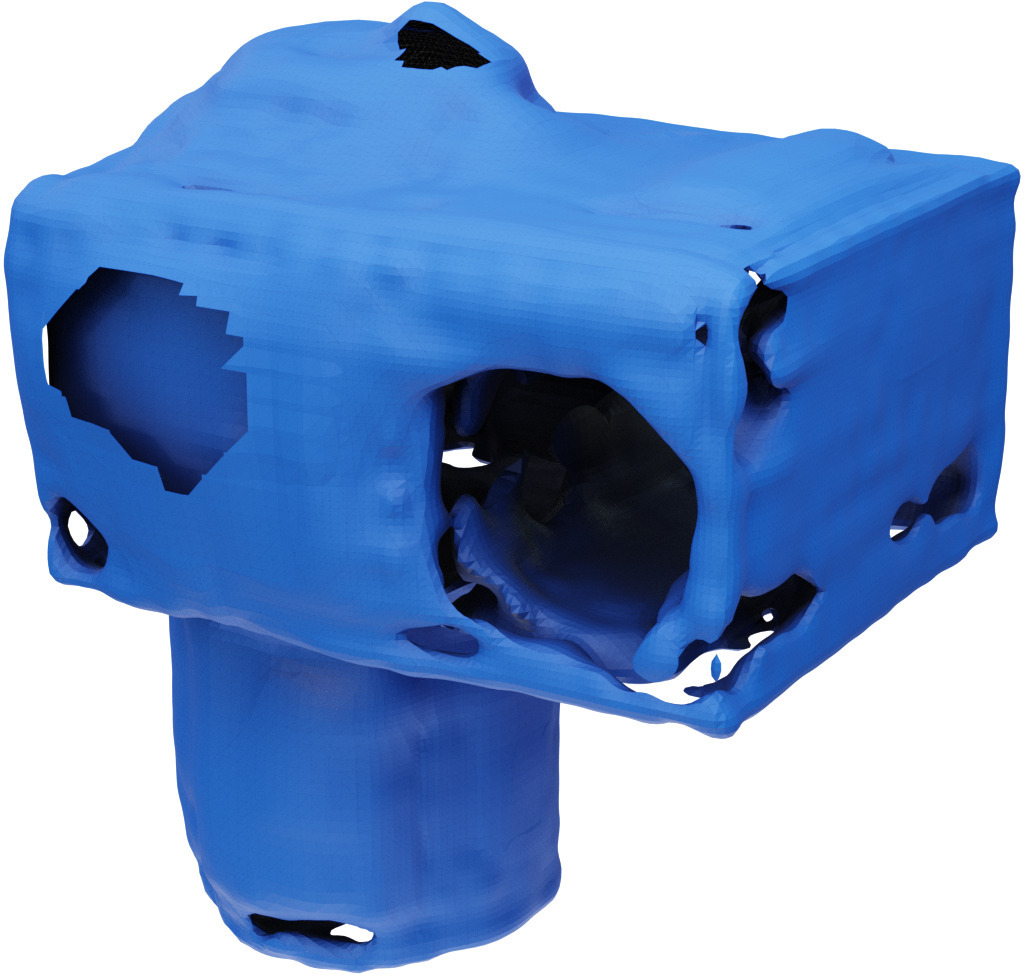}
    & \includegraphics[height=0.1\textheight,keepaspectratio]{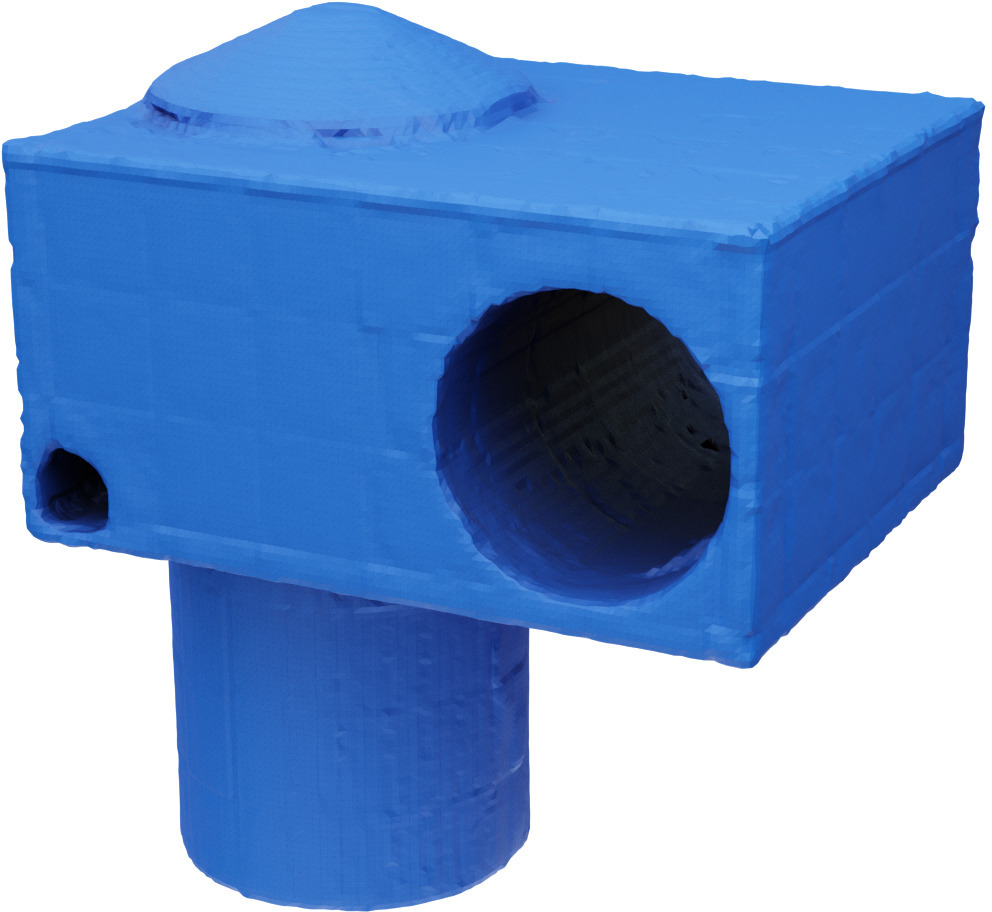}
    & \includegraphics[height=0.1\textheight,keepaspectratio]{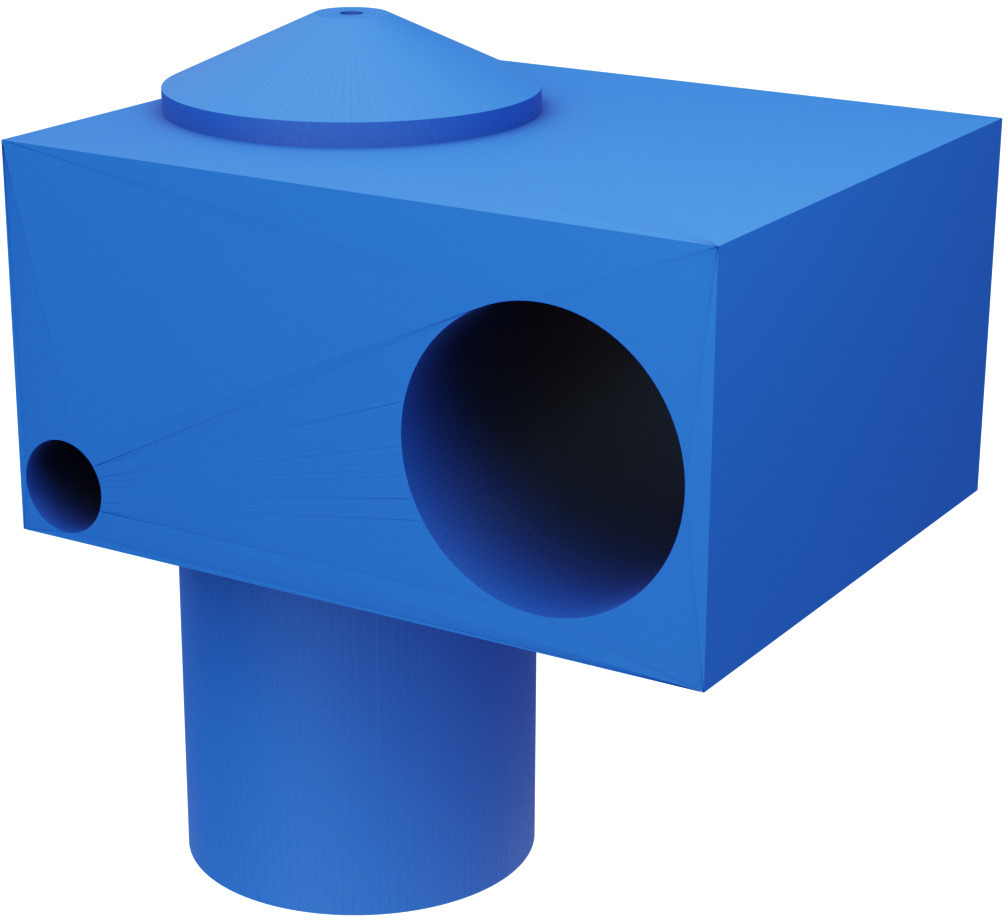} %
    \\
    \midrule
    \multirow{2}{*}{\rotatebox[origin=c]{90}{\makecell{\textbf{Famous}\\mid-noise}}}
    & \includegraphics[height=0.1\textheight,keepaspectratio]{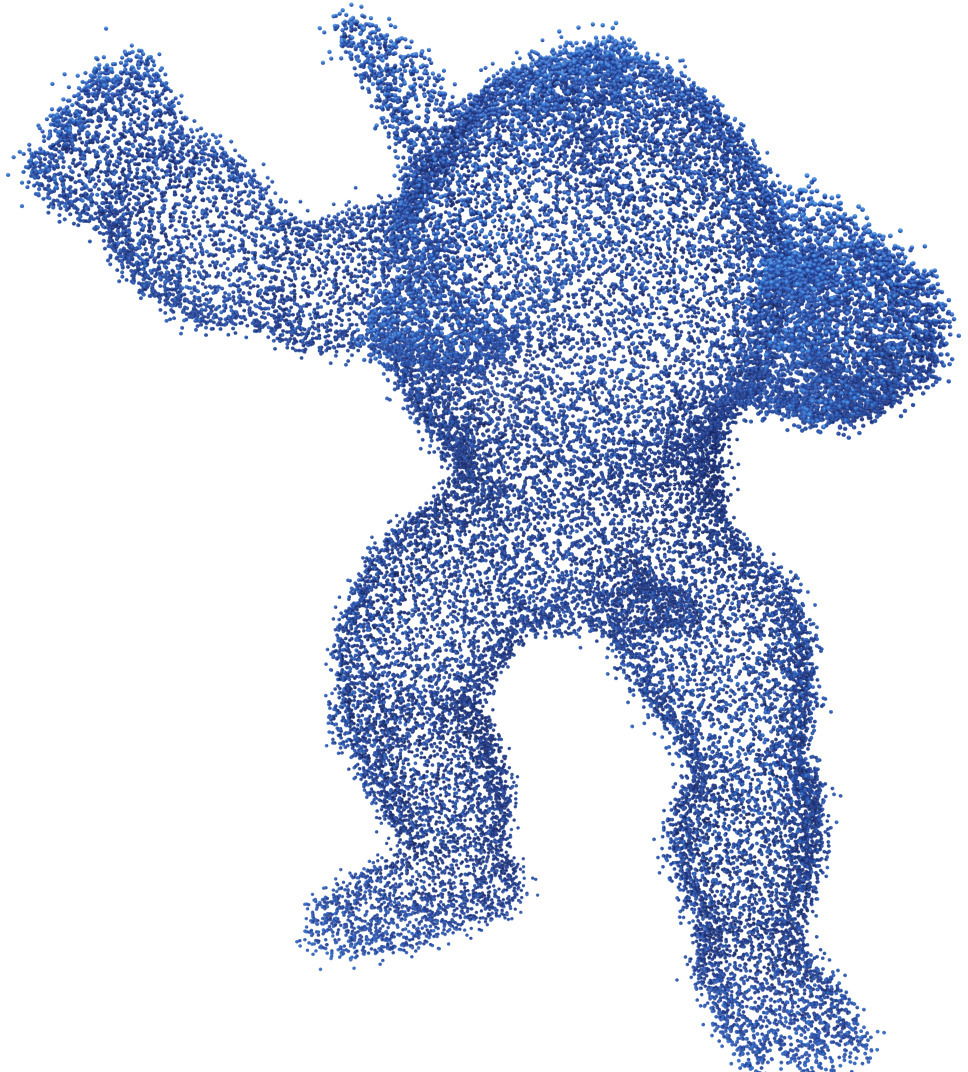}
    & \includegraphics[height=0.1\textheight,keepaspectratio]{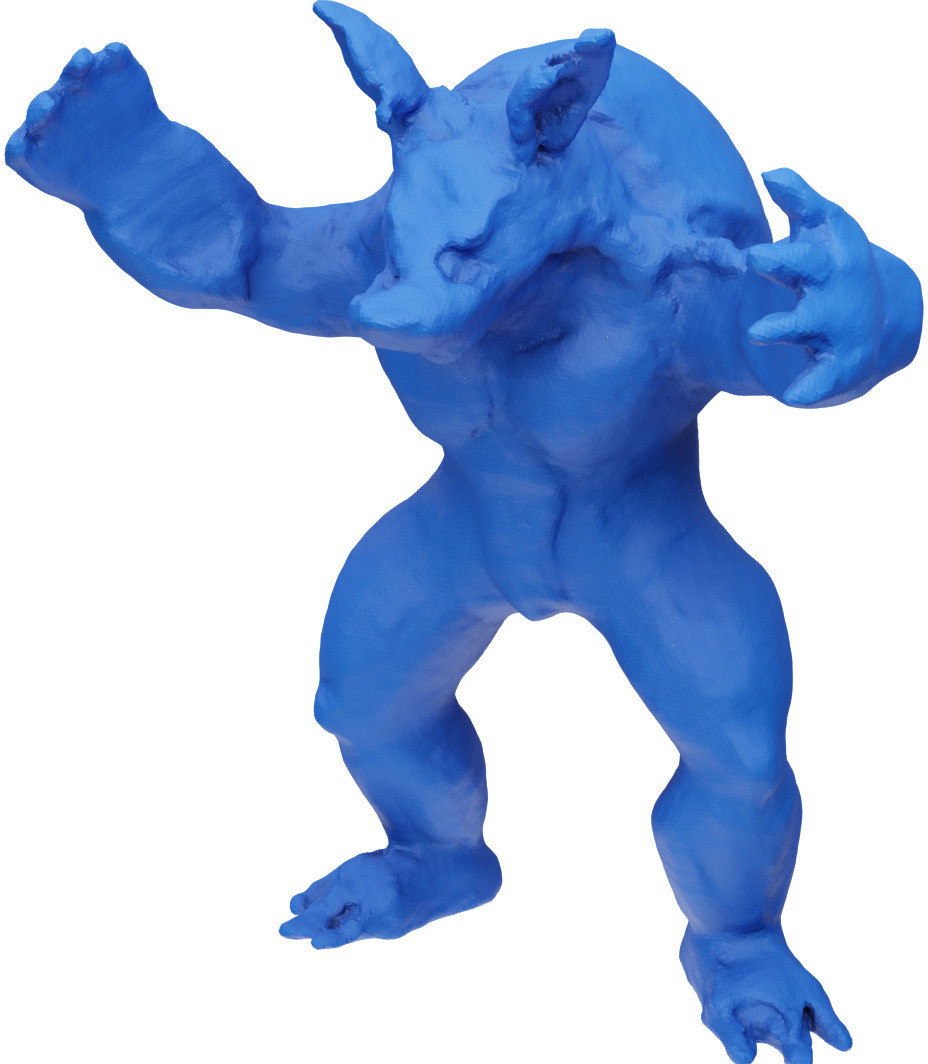}
    & \includegraphics[height=0.1\textheight,keepaspectratio]{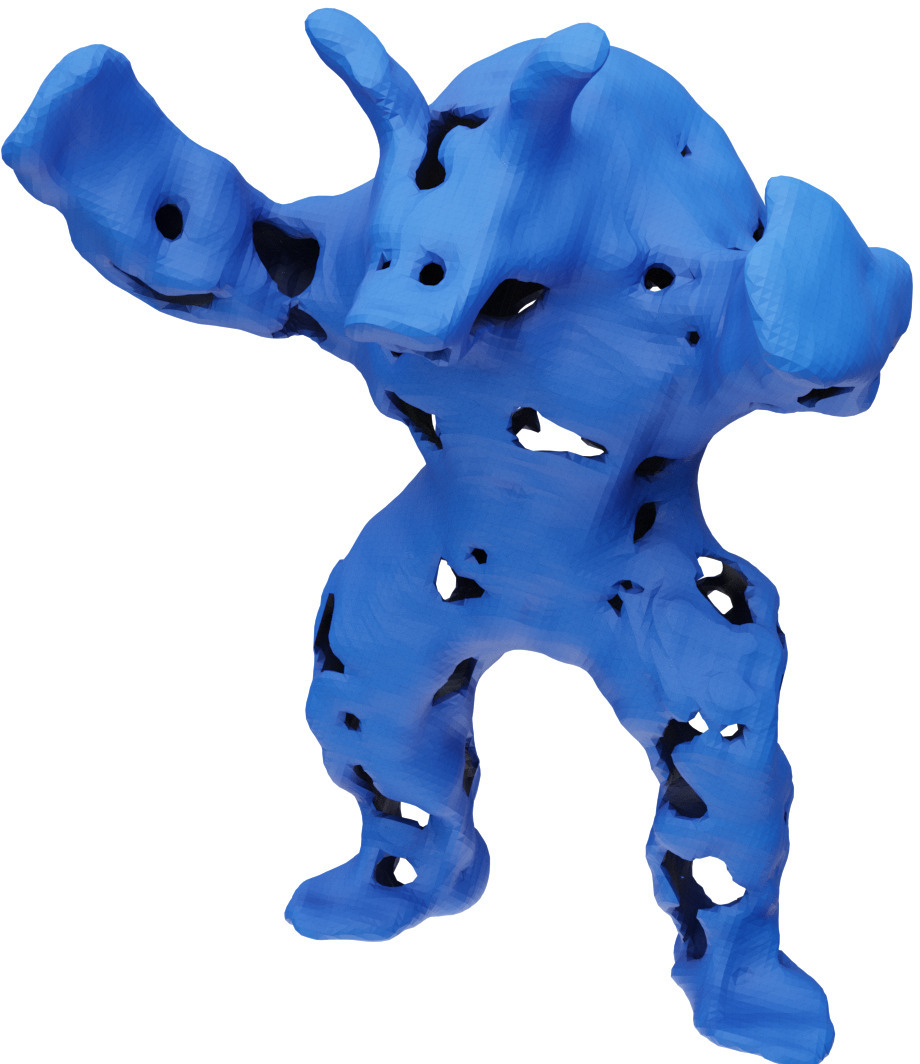}
    & \includegraphics[height=0.1\textheight,keepaspectratio]{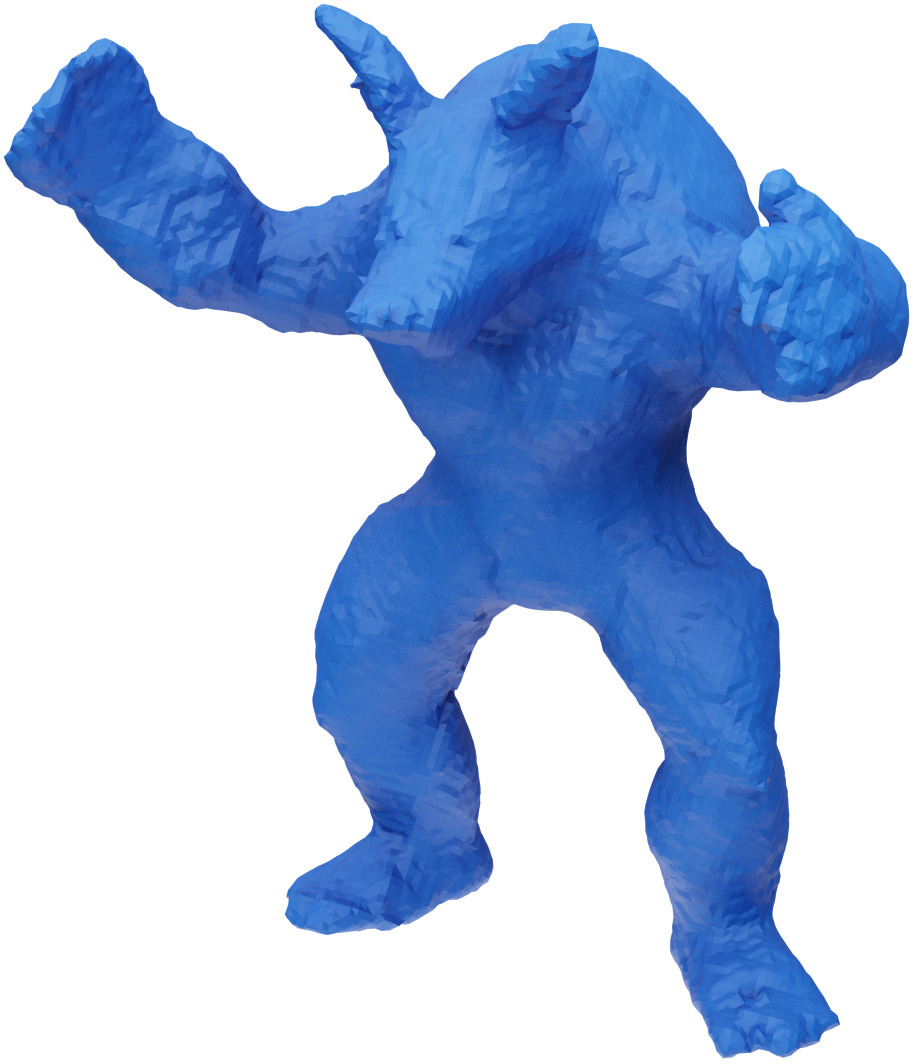}
    & \includegraphics[height=0.1\textheight,keepaspectratio]{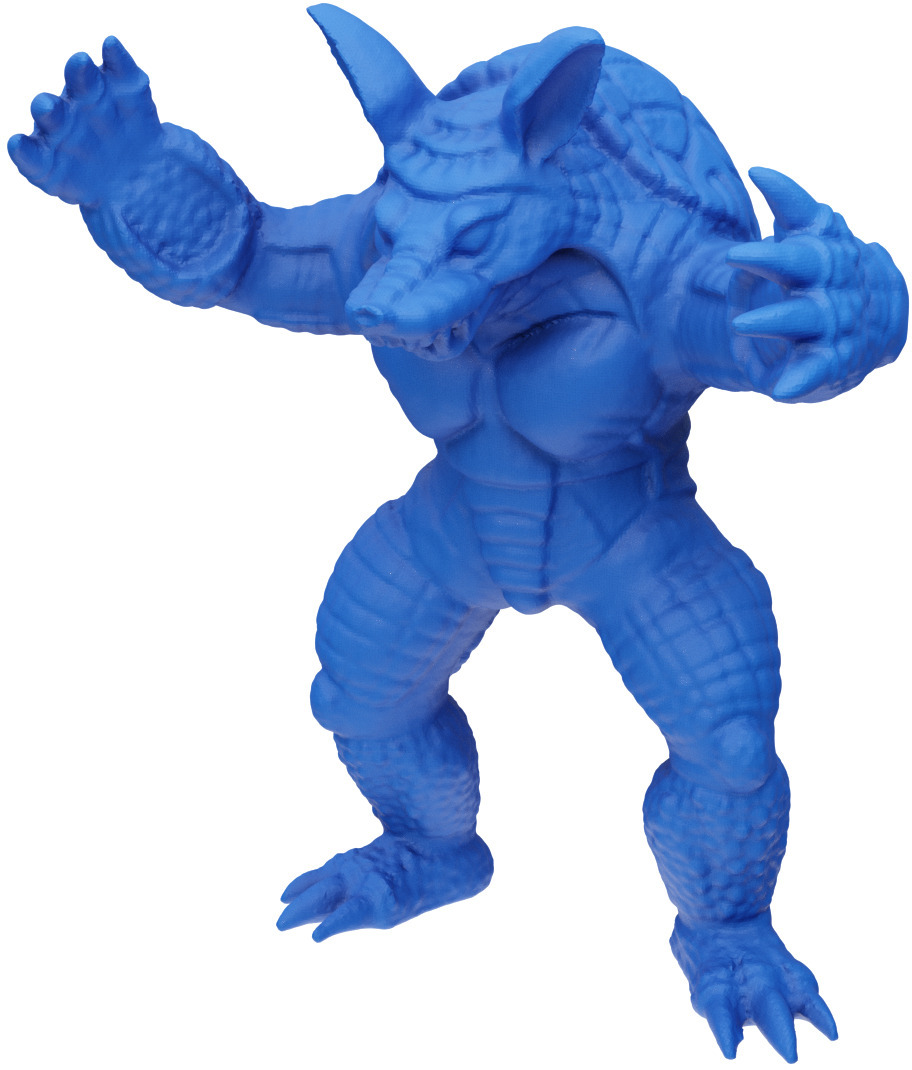} %
    \\
    & \includegraphics[height=0.1\textheight,keepaspectratio]{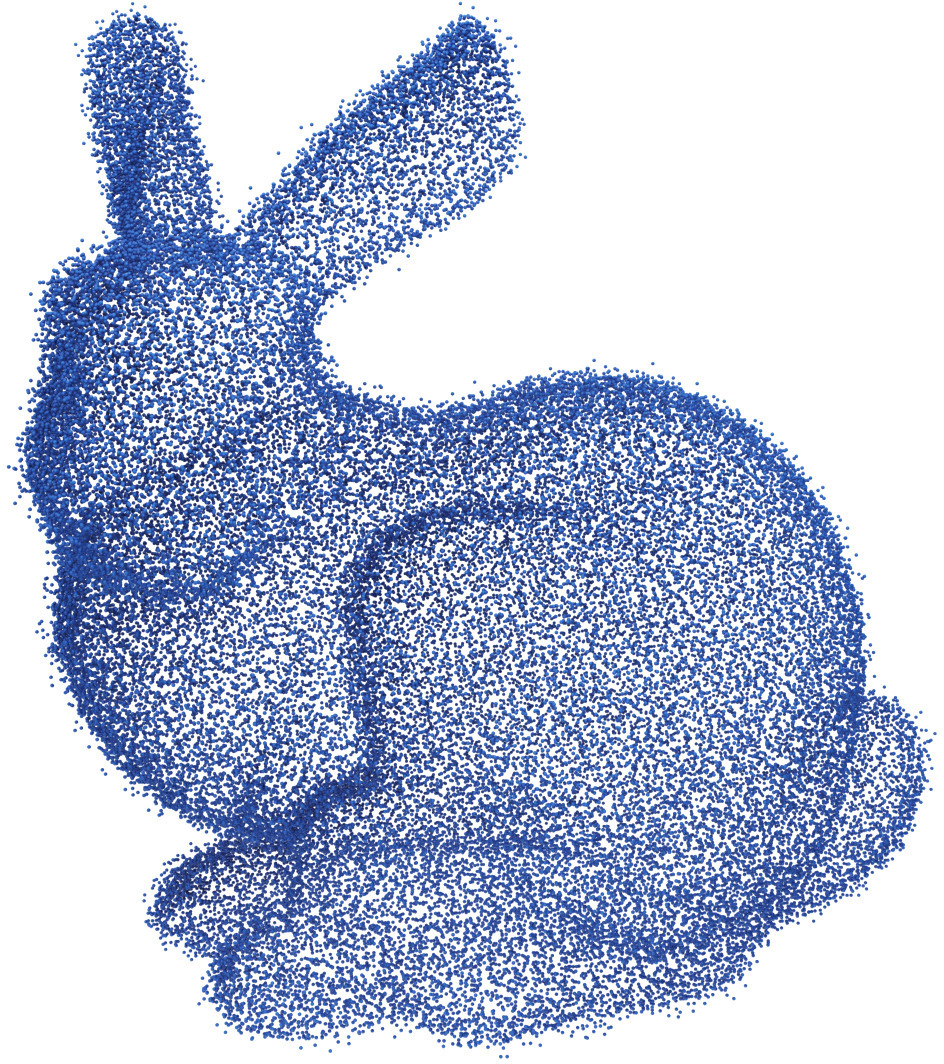}
    & \includegraphics[height=0.1\textheight,keepaspectratio]{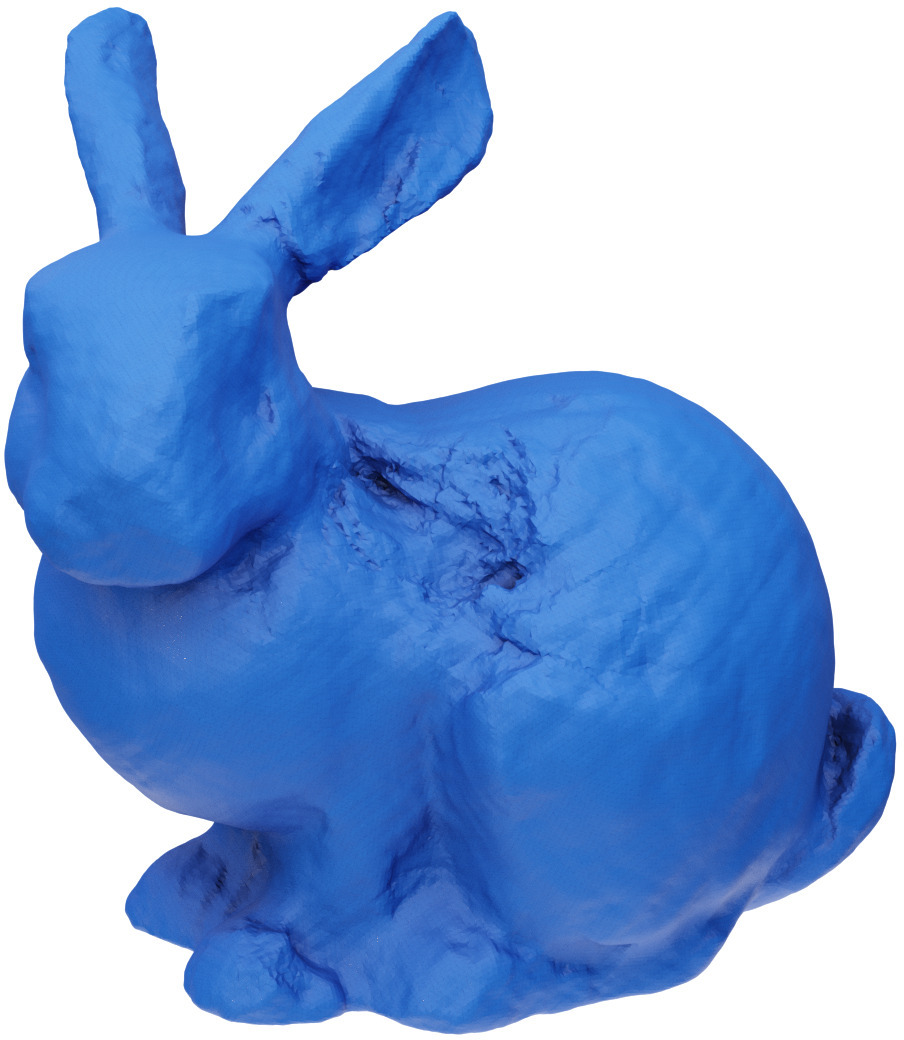}
    & \includegraphics[height=0.1\textheight,keepaspectratio]{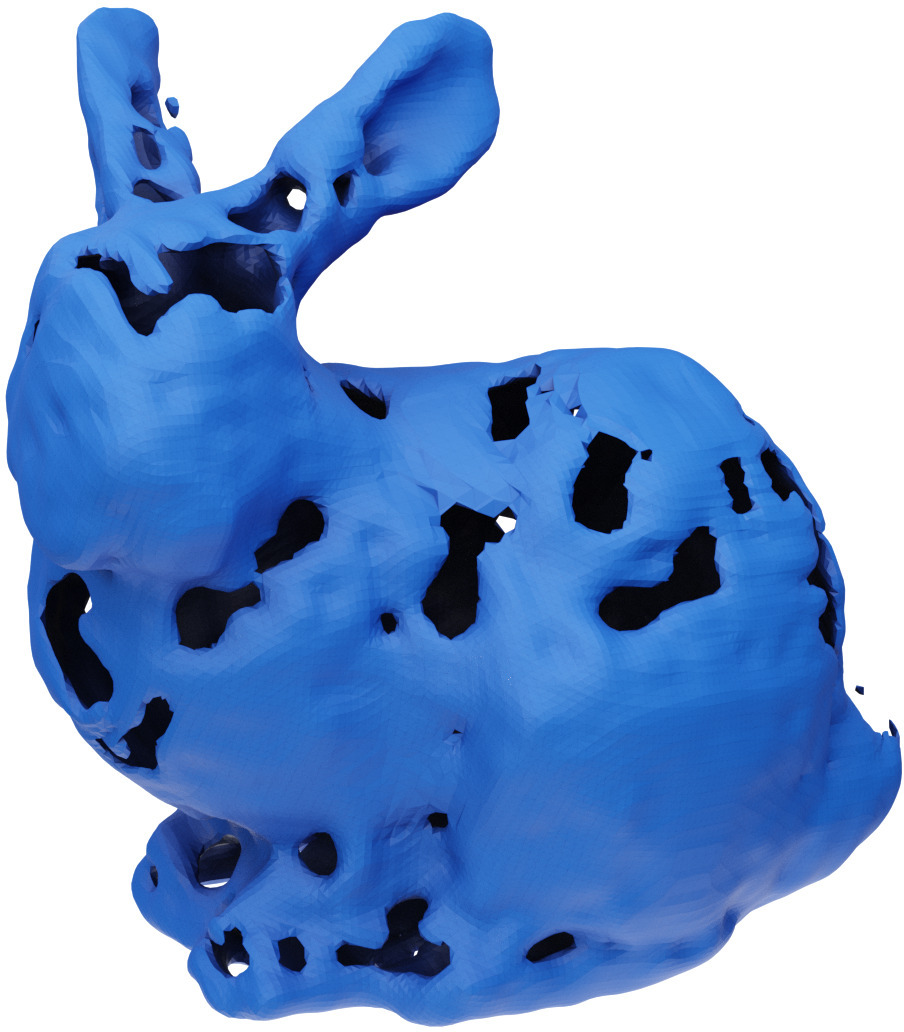}
    & \includegraphics[height=0.1\textheight,keepaspectratio]{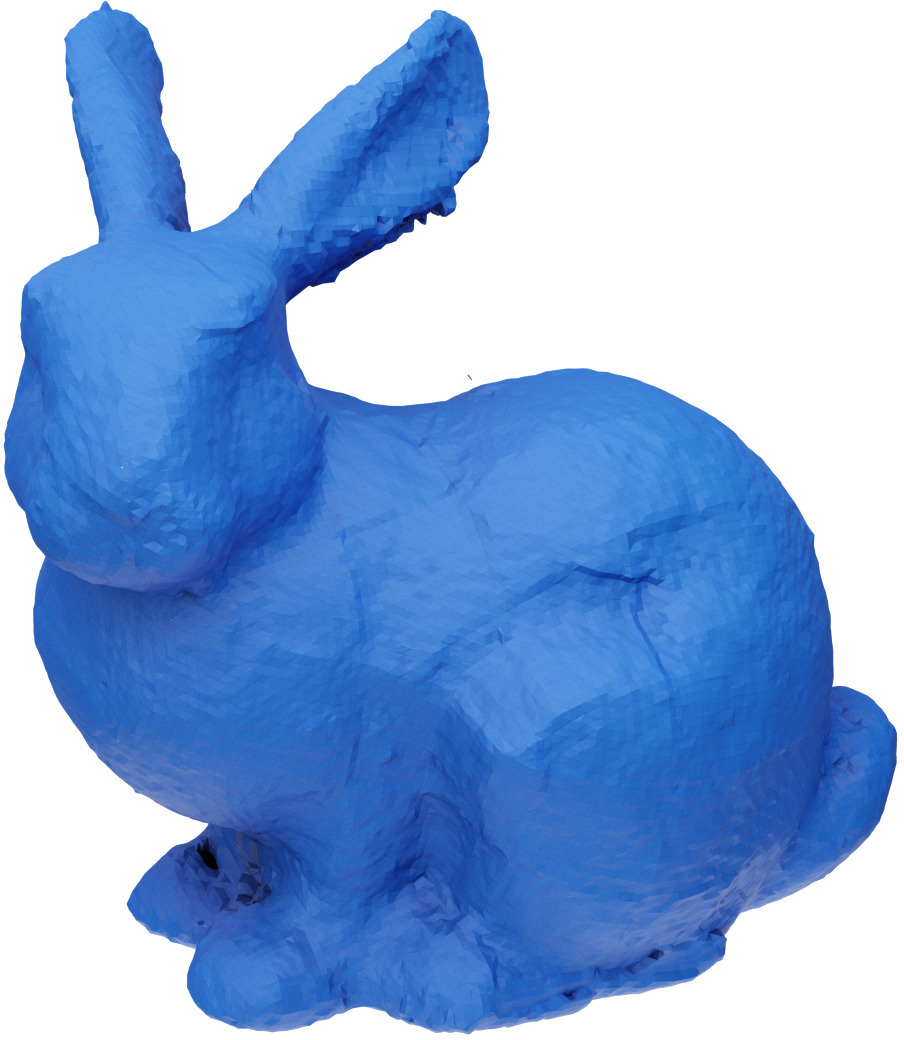}
    & \includegraphics[height=0.1\textheight,keepaspectratio]{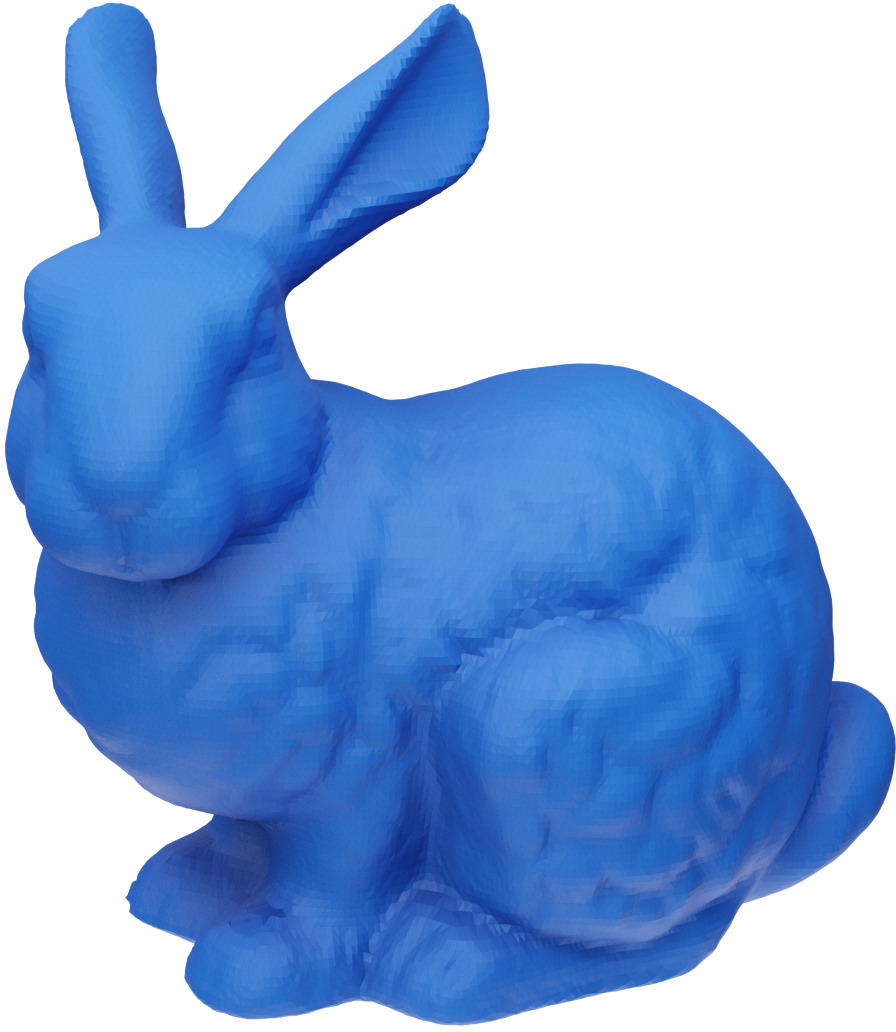} %
    \\
    \midrule
    \multirow{2}{*}{\rotatebox[origin=c]{90}{\makecell{\textbf{Thingi10K}\\sparse}}}
    & \includegraphics[height=0.1\textheight,keepaspectratio]{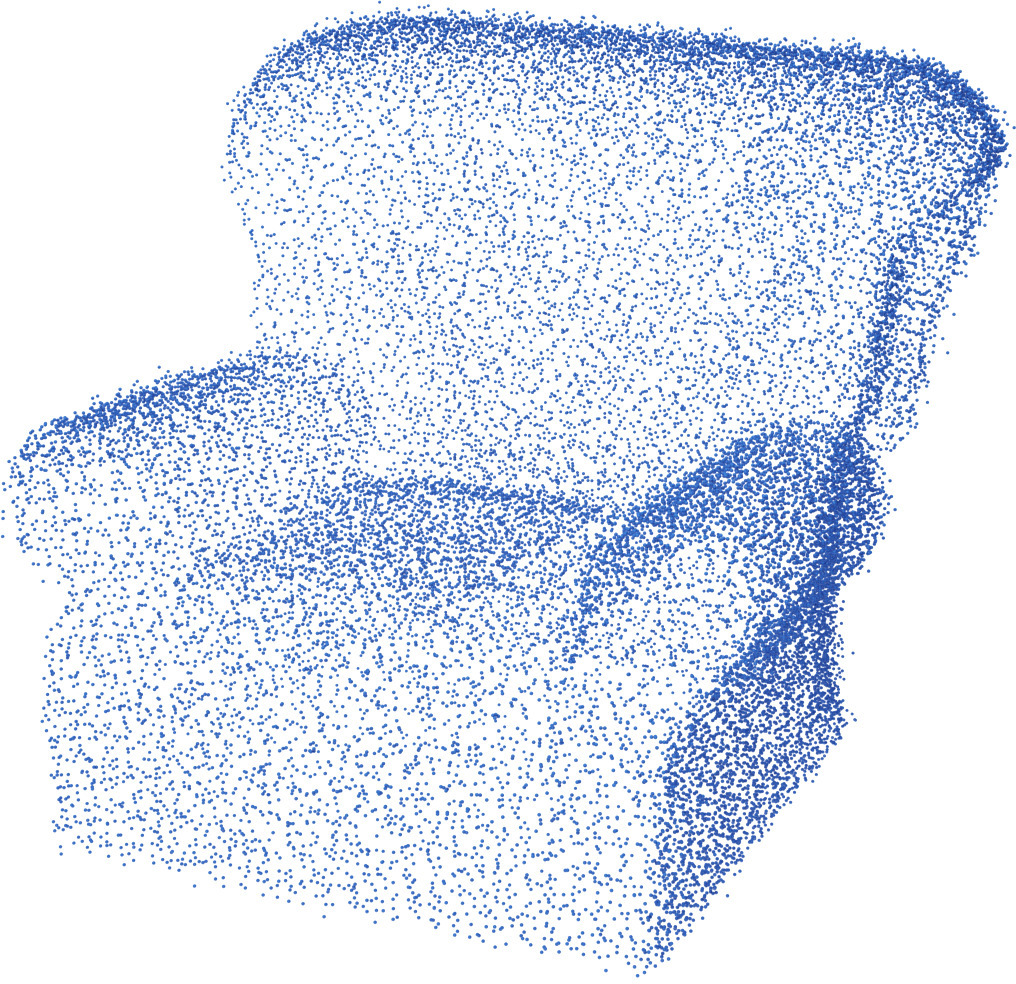}
    & \includegraphics[height=0.1\textheight,keepaspectratio]{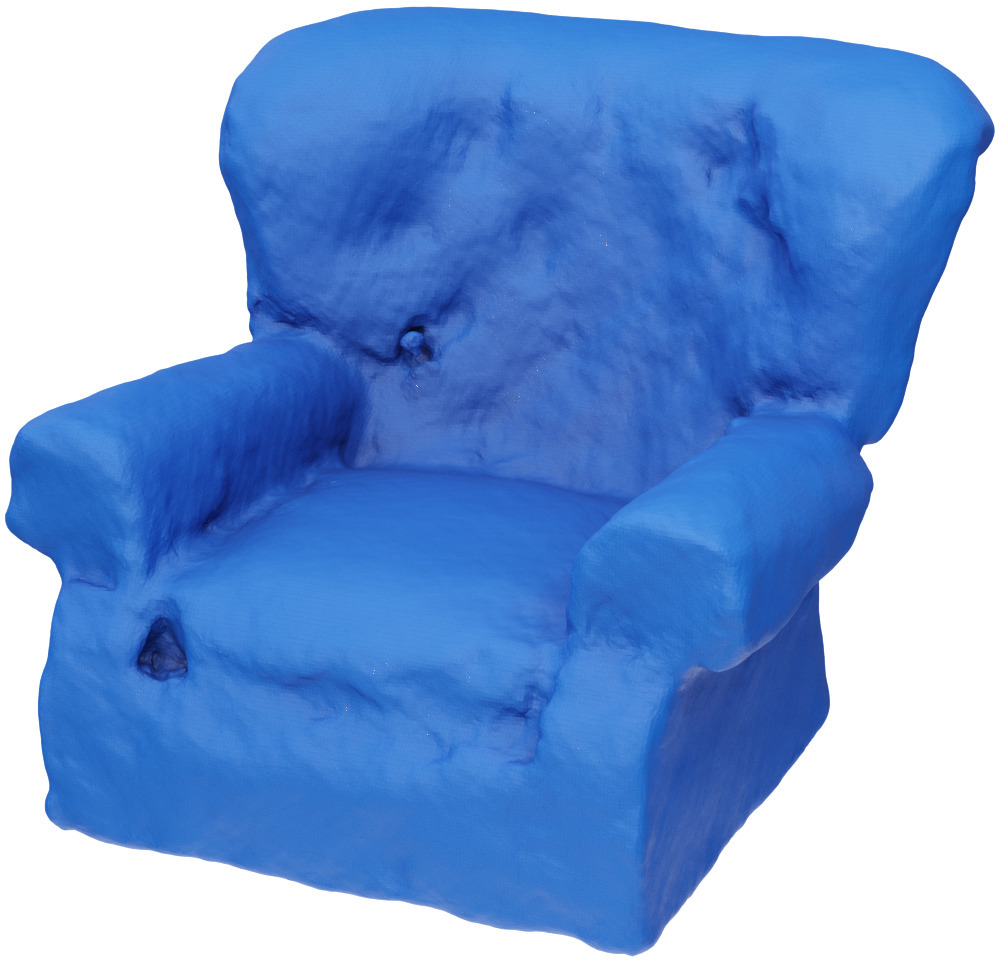}
    & \includegraphics[height=0.1\textheight,keepaspectratio]{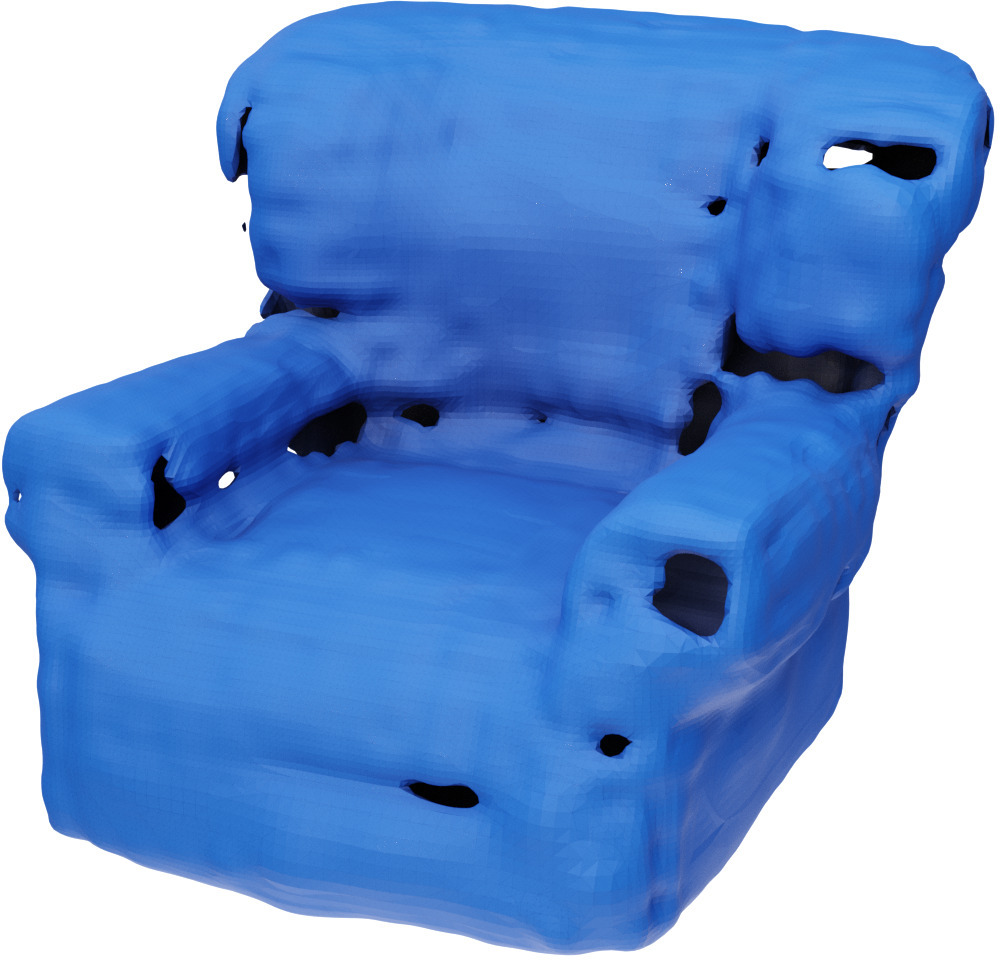}
    & \includegraphics[height=0.1\textheight,keepaspectratio]{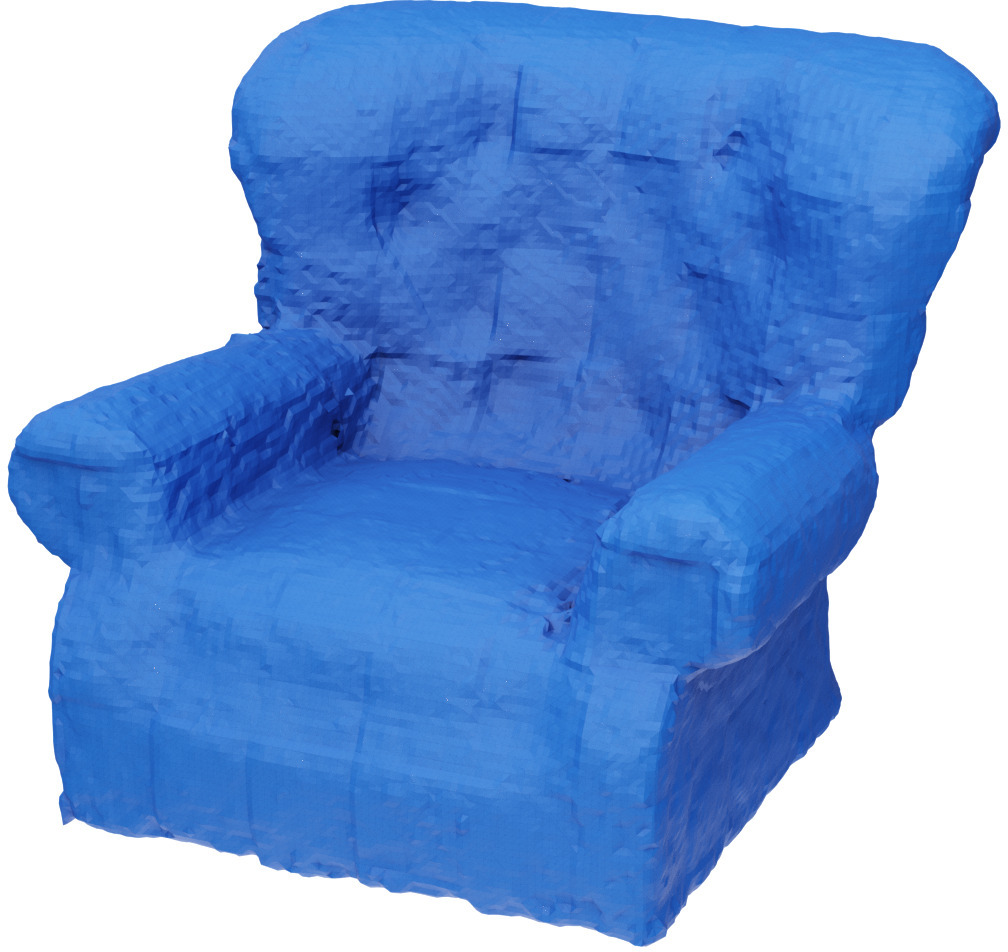}
    & \includegraphics[height=0.1\textheight,keepaspectratio]{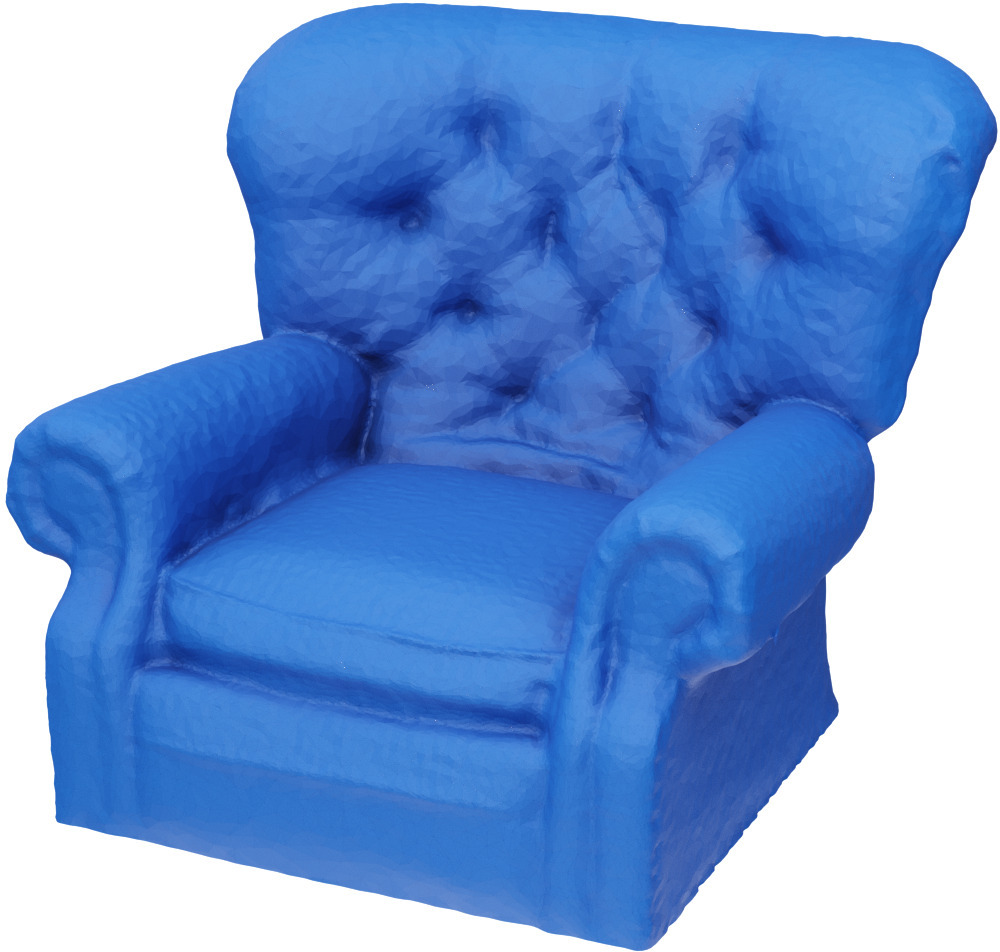} %
    \\
     
    & \includegraphics[height=0.1\textheight,keepaspectratio]{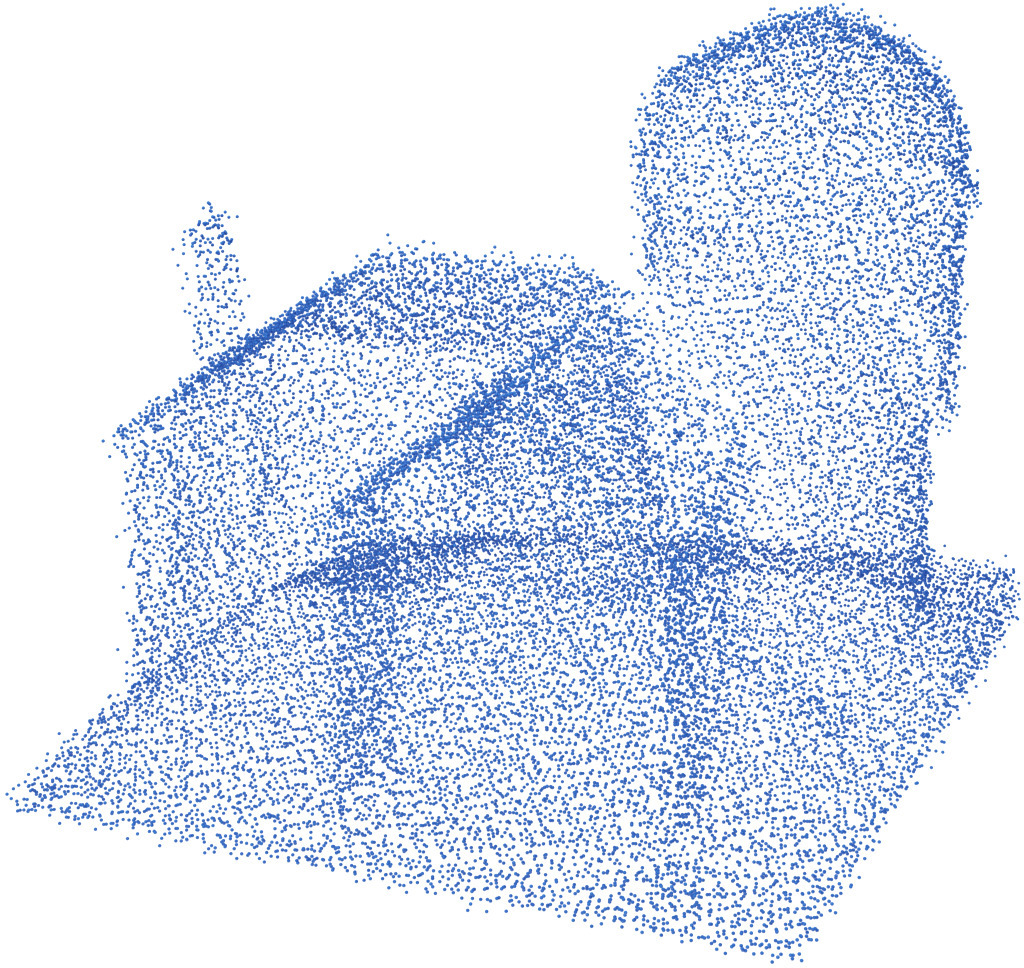}
    & \includegraphics[height=0.1\textheight,keepaspectratio]{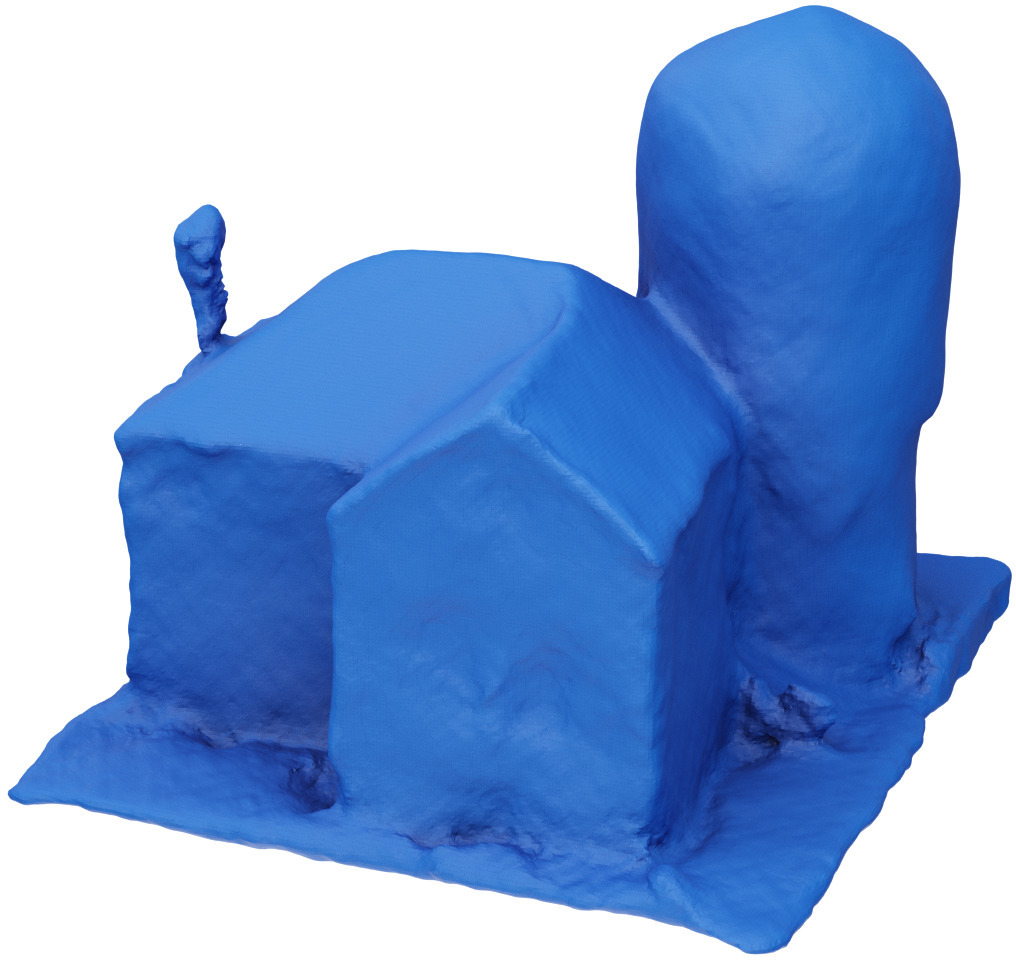}
    & \includegraphics[height=0.1\textheight,keepaspectratio]{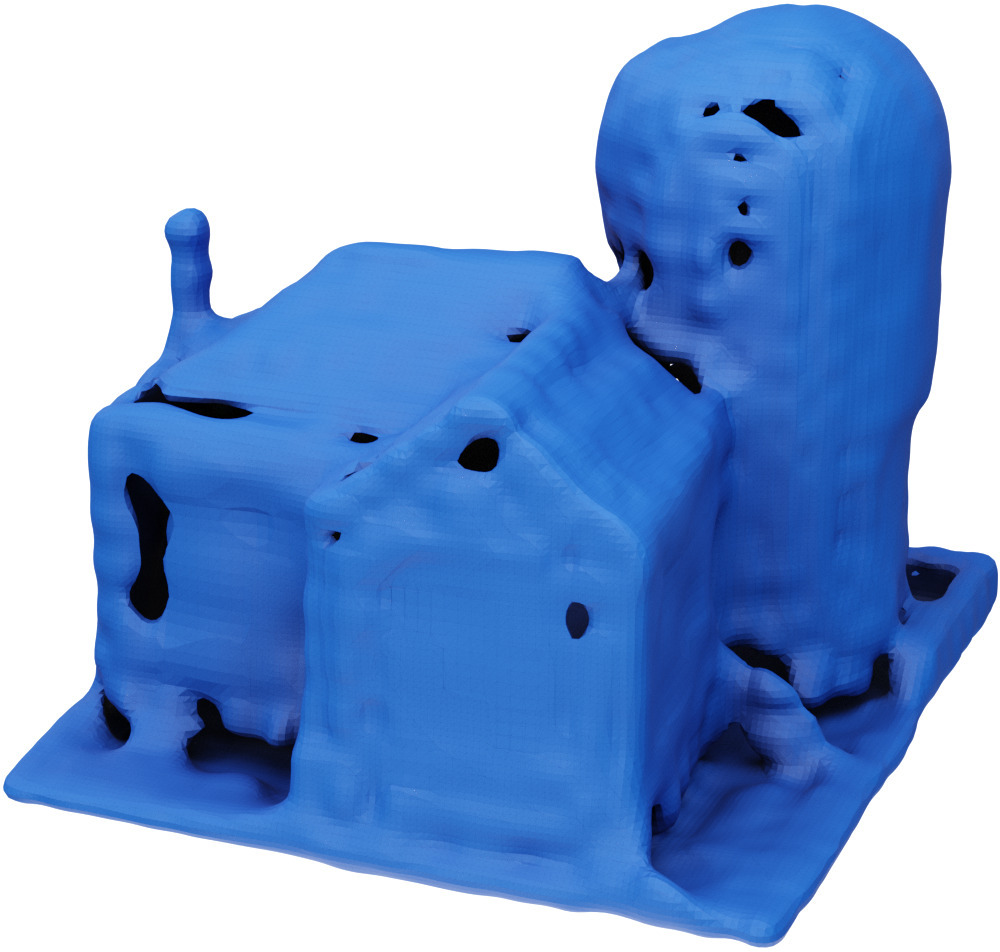}
    & \includegraphics[height=0.1\textheight,keepaspectratio]{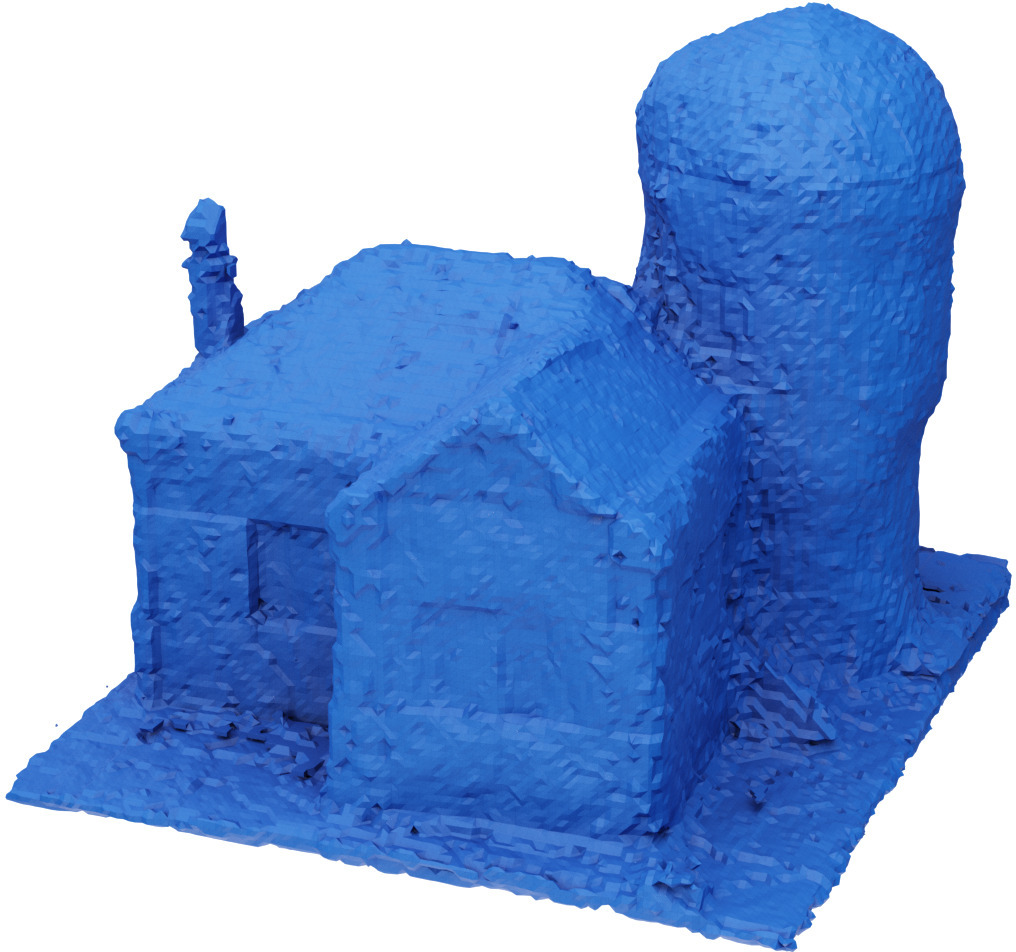}
    & \includegraphics[height=0.1\textheight,keepaspectratio]{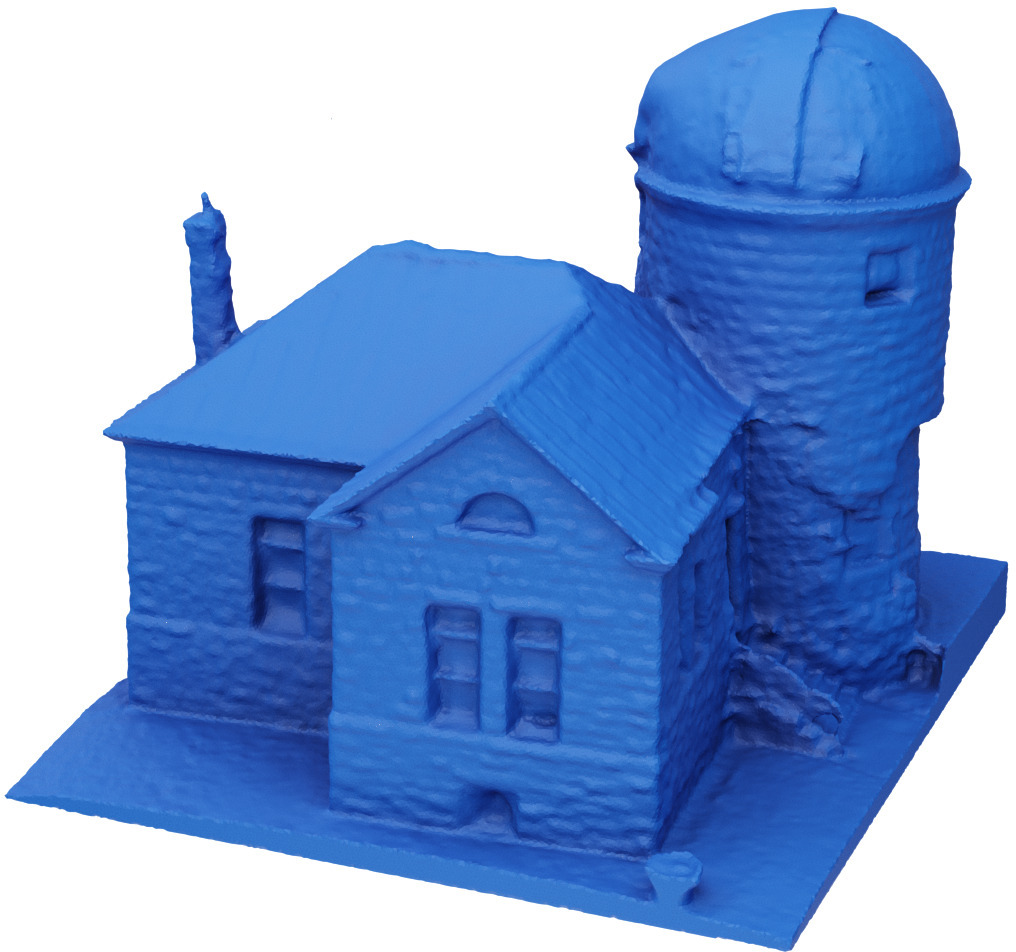} %
    \\
    \midrule
    & \textbf{Point Cloud} & \textbf{POCO} & \textbf{NKSR} & \textbf{SurfR} & \textbf{Ground Truth}
    \end{tabular}}
    \caption{\textbf{Qualitative Baseline Comparison}: the results of the proposed method are qualitatively on par with the baselines, or even better, particularly on the Famous and Thingi10K datasets, with different noise/density levels. \textbf{SurfR} preserves more detail even in sparsely sampled surfaces.}
    \label{fig:qual_comp}
\end{figure*}

\begin{figure*}[!htb]
    \setlength{\tabcolsep}{10pt}
    \centering
    \resizebox{0.8\linewidth}{!}{\begin{tabular}{>{\kern-\tabcolsep}c<{\kern-\tabcolsep} c c >{\kern-\tabcolsep}c<{\kern-\tabcolsep}}
    \includegraphics[height=0.105\textheight,keepaspectratio]{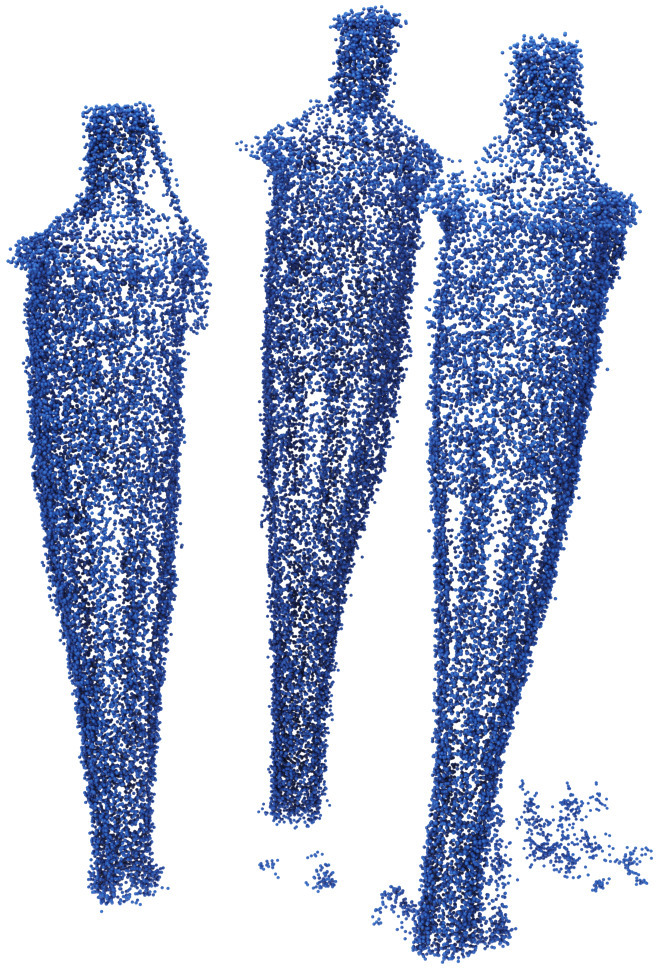}
    & \includegraphics[height=0.105\textheight,keepaspectratio]{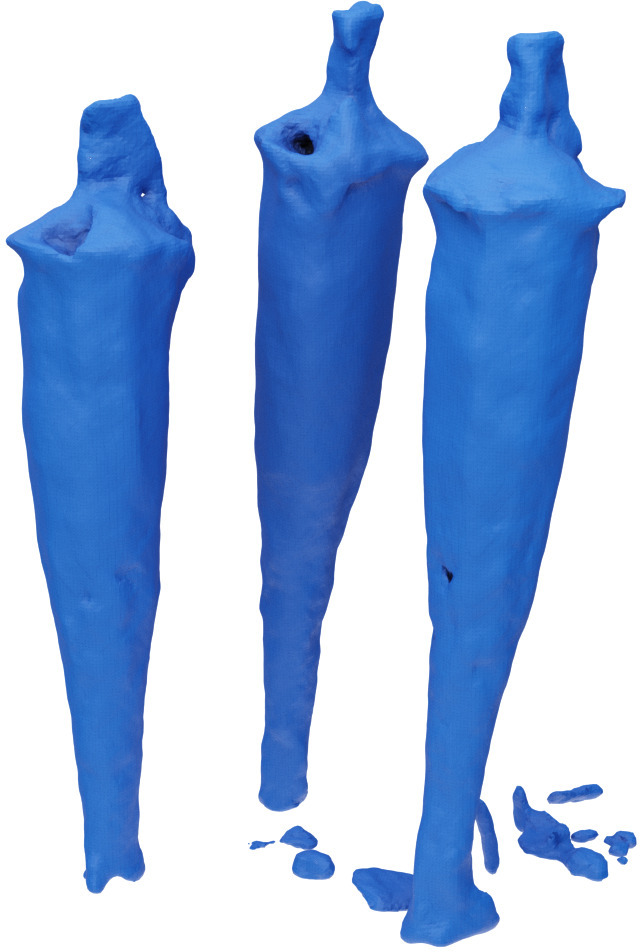}
    & \includegraphics[height=0.105\textheight,keepaspectratio]{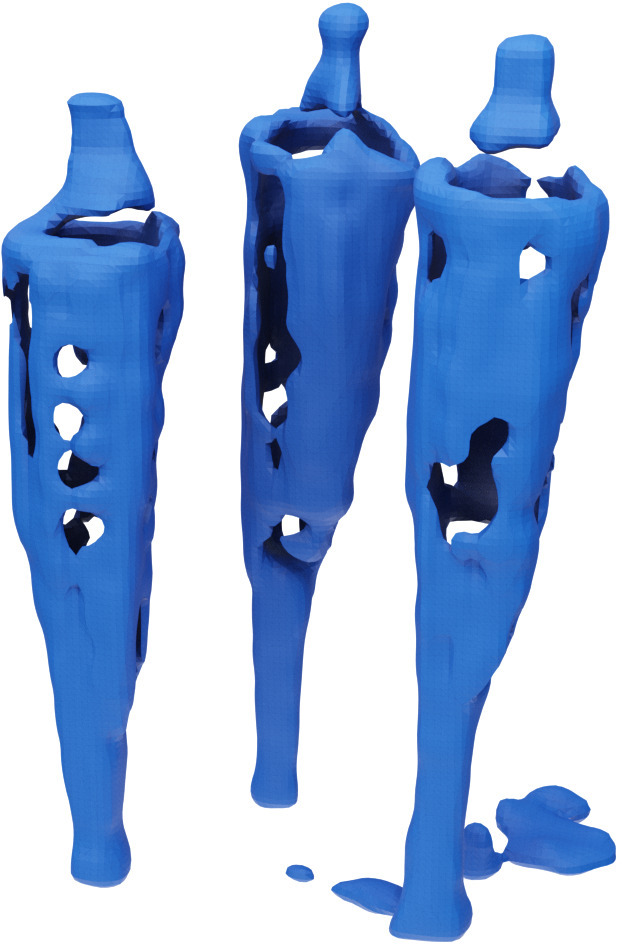}
    & \includegraphics[height=0.105\textheight,keepaspectratio]{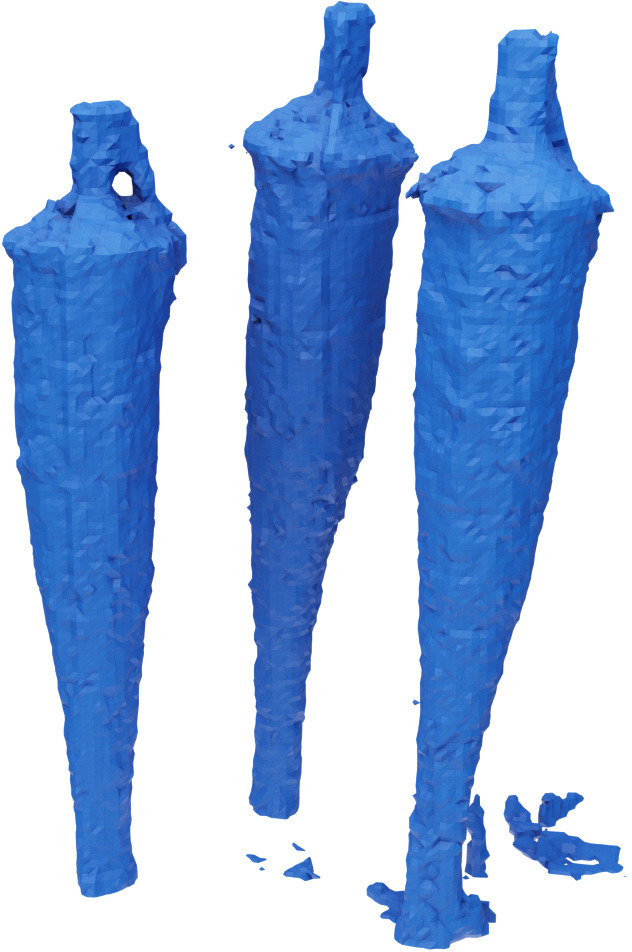} \\

    \includegraphics[height=0.105\textheight,keepaspectratio]{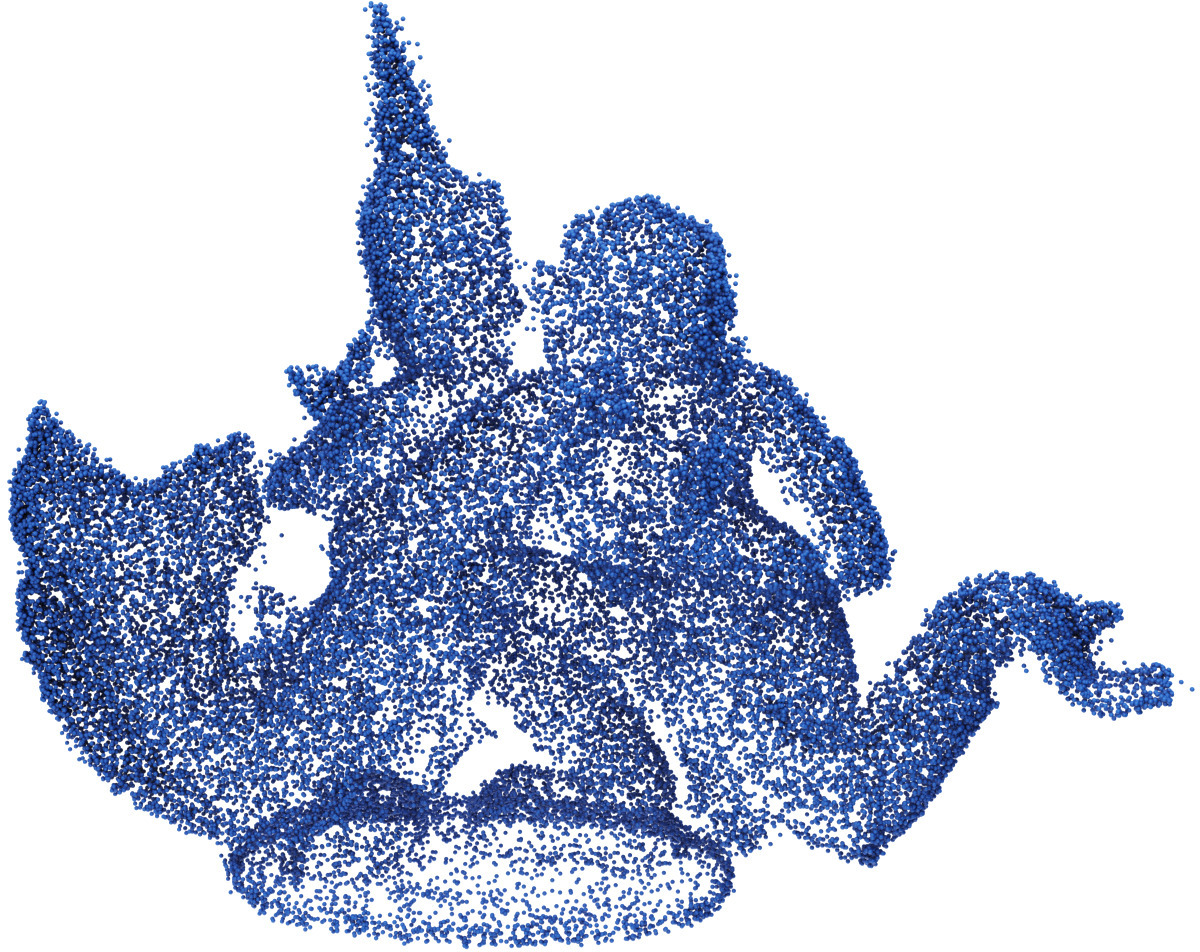}
    & \includegraphics[height=0.105\textheight,keepaspectratio]{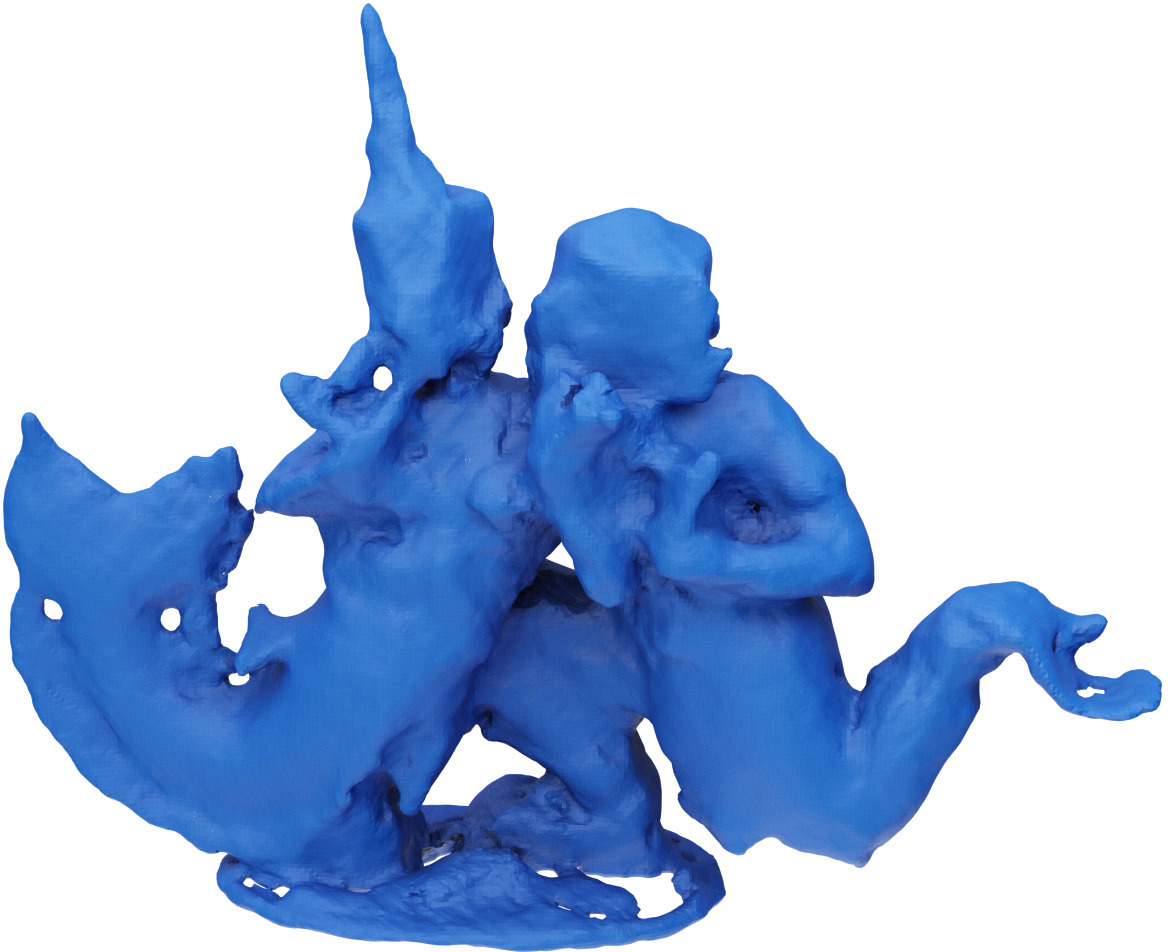}
    & \includegraphics[height=0.105\textheight,keepaspectratio]{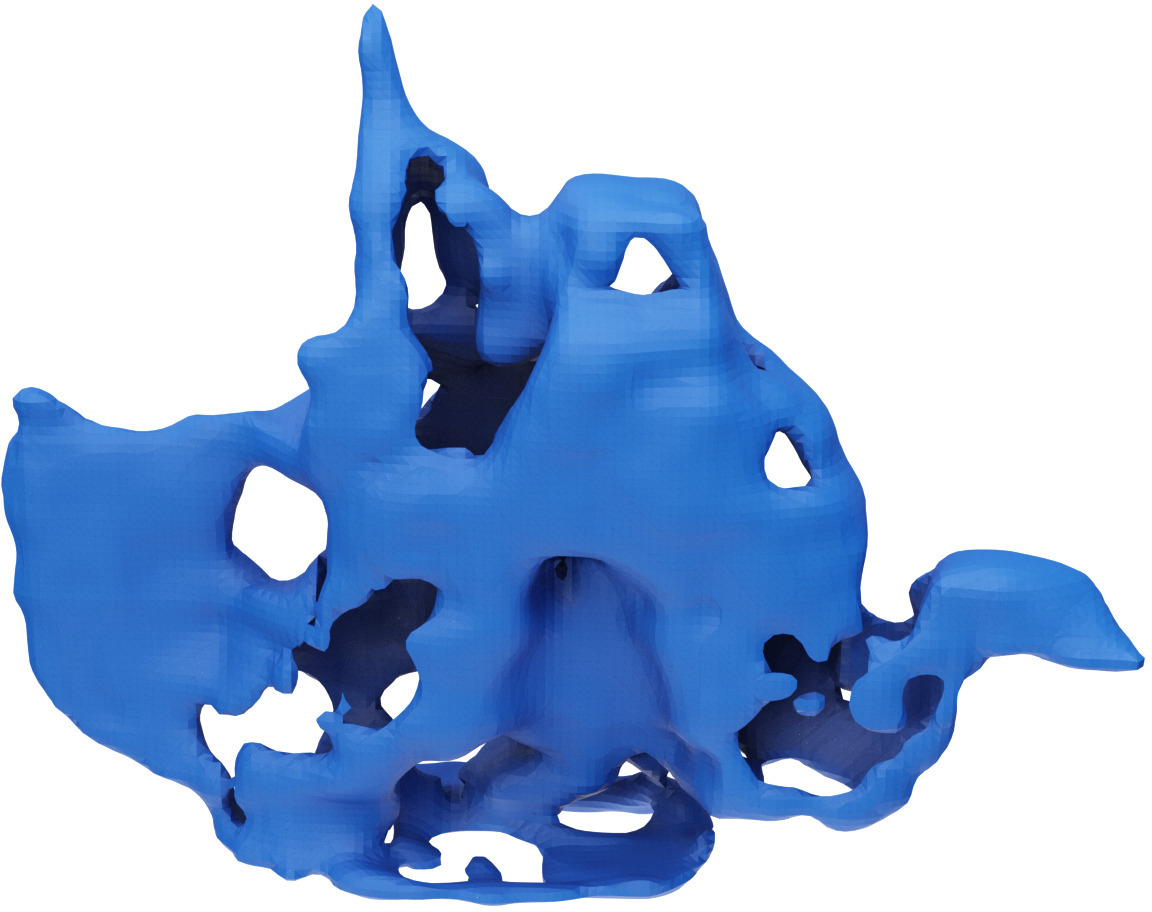}
    & \includegraphics[height=0.105\textheight,keepaspectratio]{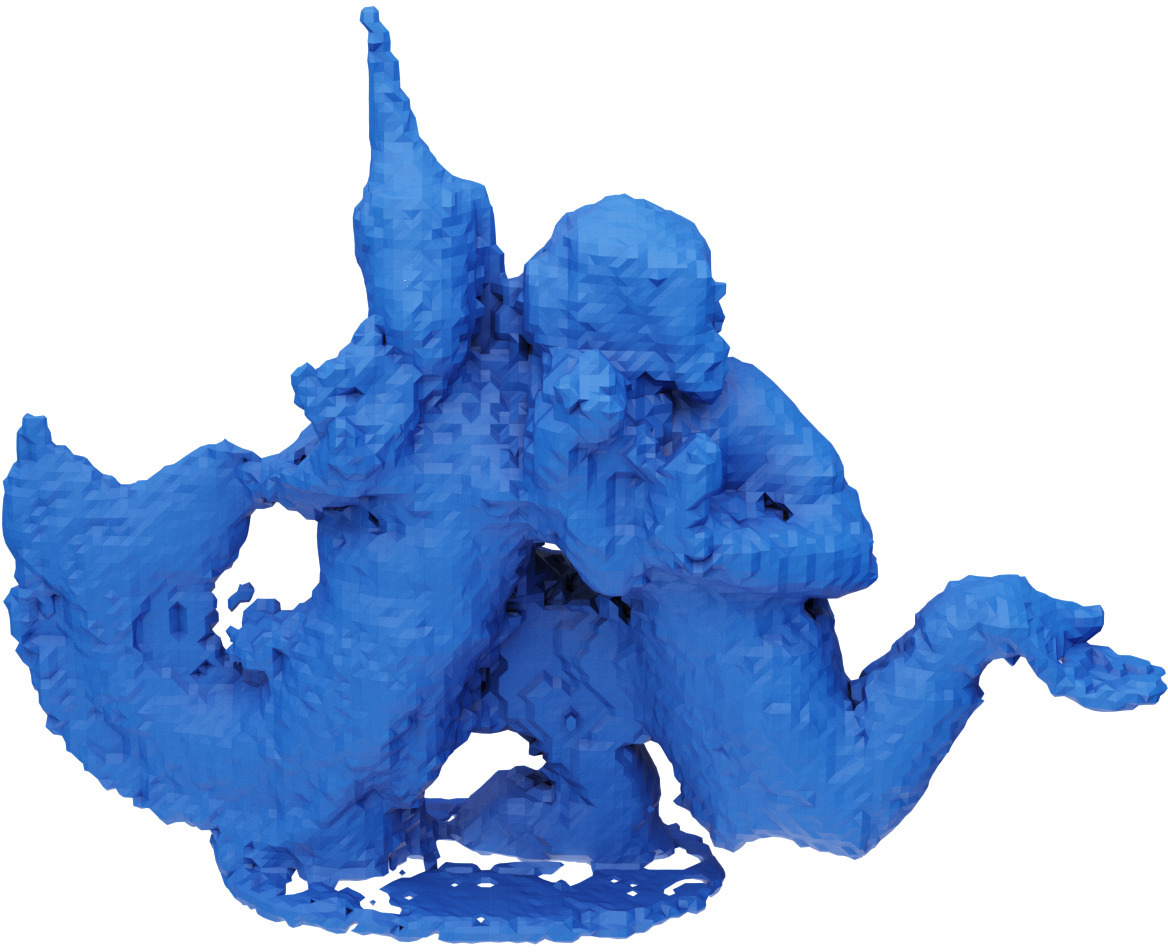} \\
    \textbf{Point Cloud} & \textbf{POCO} & \textbf{NKSR} & \textbf{SurfR}
    \end{tabular}}
    \caption{{\bf Reconstruction of Real-world Objects}: SurfR generalizes well, even in noise and variable sampling density. The proposed method shows it can handle real-world objects taken from SfM methods.}
    \label{fig:qual_comp_real}
\end{figure*}

\begin{table*}[t]
    \caption{\textbf{Quantitative Baseline Comparison}: Chamfer distance between the reconstructed and ground truth meshes (lower is better) and normal consistency measuring the accuracy of the shape normals (higher is better).}
    \centering
    \setlength{\tabcolsep}{5pt}
    \renewcommand{\arraystretch}{1.2}
    \resizebox{0.94\linewidth}{!}{%
    \begin{tabular}{>{\kern-\tabcolsep}l<{\kern-\tabcolsep} c c c c c c c c c c c c c c c c}
    \toprule
    & \multicolumn{9}{c}{\thead{Chamfer Distance $L_2$~($\downarrow$)}} &
    \multicolumn{7}{c}{\thead{Normal consistency~($\uparrow$)}}\\ 
    \cmidrule(lr){2-10} \cmidrule(lr){11-17}
    
    \multirow{2}{*}{\thead{Noise \&\\Sparsity}\hspace{5pt}} &
    \multirow{2}{*}{\thead{CON\\\cite{peng2020convolutional}}} &
    \multirow{2}{*}{\thead{P2S\\\cite{Point2Mesh}}} &
    \multirow{2}{*}{\thead{SAP\\\cite{Peng2021}}} &
    \multirow{2}{*}{\thead{IF\\\cite{chibane2020implicit}}} &
    \multirow{2}{*}{\thead{POCO\\\cite{boulch2022poco}}} &
    \multirow{2}{*}{\thead{NKSR\\\cite{huang2023neural}}} &
    \multicolumn{3}{c}{\bf SurfR} & 
    \multirow{2}{*}{\thead{CON\\\cite{peng2020convolutional}}} &
    \multirow{2}{*}{\thead{IF\\\cite{chibane2020implicit}}} &
    \multirow{2}{*}{\thead{POCO\\\cite{boulch2022poco}}} &
    \multirow{2}{*}{\thead{NKSR\\\cite{huang2023neural}}} &
    \multicolumn{3}{c}{\bf SurfR}\\  
    
    \cmidrule(lr){8-10}  \cmidrule(lr){15-17}
    \multicolumn{1}{c}{} & & & & & & & \textbf{64} & \textbf{128} & 
    \textbf{256}  & & & & & \textbf{64} & \textbf{128} & \textbf{256} \\ \midrule
    & \multicolumn{13}{c}{ABC dataset~\cite{Koch_2019_CVPR} (unseen in training)} \\

    no-noise & 2.6 & \second{1.8} & 7.6 & 2.8 & \best{1.7} & 3.0 & 2.3 & 2.1 & 2.1 & 
    \best{0.89} & 0.62 & \best{0.89} & 0.29 & \second{0.87} & \best{0.89} & \best{0.89} \\
    
    med-noise & 7.7 & \best{2.1} & 8.1 & 3.8 & \best{2.1} & 3.3 & 2.5 & \second{2.4} & \second{2.4} & 
    0.8 & 0.61 & \best{0.84} & 0.33 & 0.81 & \second{0.82} & 0.80 \\

    max-noise &  14.0 & \second{2.8} & 7.2 & 5.9 & \best{2.7} & 4.1 & 3.3 &  3.2 & 3.1 & 
    0.74 & 0.57 & 0.75 & 0.4 & \best{0.78} & \second{0.75} & 0.71 \\ \midrule

    & \multicolumn{13}{c}{Famous dataset (a collection of 22 popular meshes)} \\
    no-noise & 2.4 & \best{1.4} & 9.0 & 2.2 & \best{1.4} & 2.7 & 1.8 & \second{1.5} & \best{1.4} & 
    \second{0.84} & 0.63 & 0.78 & 0.18 & 0.82 & \best{0.86} & \best{0.86} \\

    med-noise & 3.3 & \best{1.5} & 8.8 & 2.3 & 1.7 & 2.7 & 1.9 &  1.7 & \second{1.6} & 
    \best{0.83} & 0.63 & 0.73 & 0.17 & \second{0.81} & \best{0.83} & \second{0.81} \\

    max-noise & 12.8 & \best{2.5} & 7.0 & 5.2 & \second{2.9} & 3.9 & 3.3 &  3.1 & 3.0 & 
    \second{0.7} & 0.56 & 0.55 & 0.26 & \best{0.72} & \second{0.7} & 0.68 \\

    sparse & 3.2 & \best{1.9} & 10.4 & 2.6 & \second{2.0} & 3.2 & 2.4 & 2.1 & 2.1 & 
    \best{0.81} & 0.63 & 0.67 & 0.18 & 0.77 & \second{0.79} & 0.77 \\

    dense & 3.7 & \best{1.3} & 7.8 & 2.4 & \second{1.5} & 2.3 & 1.8 & 1.6 & \second{1.5} & 
    \best{0.84} & 0.64 & 0.76 & 0.19 & 0.82 & \best{0.84} & \second{0.83} \\ \midrule

    & \multicolumn{13}{c}{Thingi10k dataset~\cite{Thingi10K}} \\
    no-noise & 1.9 & \best{1.4} & 8.4 & 2.1 & \best{1.4} & 2.7 & 1.8 & \second{1.5} & \best{1.4} & 
    \best{0.92} & 0.65 & 0.92 & 0.22 & 0.89 & \best{0.92} & \second{0.91} \\

    med-noise & 3.1 & \best{1.5} & 8.2 & 2.4 & \best{1.5} & 2.7 & 1.9 & \second{1.6} & \best{1.5} & 
    \best{0.9} & 0.65 & \best{0.9} & 0.21 & 0.88 & \second{0.89} & 0.87 \\

    max-noise & 12.8 & 2.6 & 6.9 & 5.5 & 2.7 & 3.9 & 3.3 & 3.1 & 2.8 & 
    \second{0.76} & 0.58 & 0.71 & 0.3 & \best{0.79} & \second{0.76} & 0.73 \\

    sparse & 3.1 & \best{2.1}  & 10.1 & 2.8 & \best{2.1} & 3.3 & 2.7 & \second{2.2} & 2.4 & 
    \best{0.88} & 0.64 & 0.81 & 0.21 & 0.82 & \second{0.84} & 0.82 \\

    dense & 3.5 & \best{1.4} & 7.1 & 2.4 & \best{1.4} & 2.3 & 1.8 & \second{1.5} & \best{1.4} & 
    \best{0.91} & 0.65 & \best{0.91} & 0.2 & \second{0.89} & \best{0.91} & \second{0.89}\\
    
    \rowcolor{Gray!50} Avg.  & 5.7 & \best{1.9} & 8.2 & 3.2 & \best{1.9} & 3.1 & 2.4 & \second{2.1} &  \second{2.1} & \second{0.83} & \best{0.84} & 0.62 & 0.79 & 0.24 & \second{0.83} & 0.81 \\ \toprule

    \end{tabular}}
    \label{tab:comp_quant}
\end{table*}

\paperpar{Quantitative results}
We report the chamfer distance (multiplied by 100) and normal consistency~\cite{mescheder2019occupancy} on the test datasets (see \cref{ss:data}) in \cref{tab:comp_quant} at \cpageref{tab:comp_quant}. 
The chamfer distance measures the geometric quality of the reconstructed surface, penalizing missing and extra geometry, whereas normal consistency penalizes surface roughness and over-smoothing. 
For a fair comparison, we consider SurfR with three different surface reconstruction resolutions: $64$, $128$, and $256$. The results show that SurfR is competitive with the current state-of-the-art at a speed-up. Considering all methods with the same resolution, SurfR is 33 times faster than P2S and 7.3 times faster than POCO (both at 256 resolution), as shown in~\cref{tab:timings} and~\cref{fig:teaser}. 
The results of no-noise, med-noise, or dense reconstruction are very competitive with those of P2S and POCO; for example, in Thingi10k, SurfR achieves about the same reconstruction results as P2S and POCO. 
The main difference is in the max-noise case, where the proposed method obtains slightly worse results than both P2S and POCO and where CON, IF-Net, and NKSR fail as well since they are constrained to 3D convolutions and the kernel for NKSR.
IF-Net creates holes in the generated meshes, explaining the lower normal consistency. CON produces smoothed meshes at optimal 32 resolution, losing fine detail and resulting in a high chamfer distance, though its overly smoothed surfaces achieve a high normal consistency. NKSR produces many holes that, as a consequence, obtain low normal consistency and high chamfer distance. Additionally, SurfR is faster and more accurate than IF-Net, CON, and NKSR while running at the same or higher resolution (CON 32, IF-Net 128, NKSR 256).
SurfR, P2S, and POCO perform better at filtering noise and preserving fine detail, corroborated by the low chamfer distance and high normal consistency.

\begin{table}[t]
\caption{{\bf Inference Time Comparison in Seconds}: For the test, we compute the average 3D reconstruction time per mesh.}
\resizebox{0.95\linewidth}{!}{\setlength{\tabcolsep}{12pt}\resizebox{.5\linewidth}{!}{%
    \begin{tabular}{c c c c c c c c}
    \toprule
    \multirow{2}{*}{\thead{CON\\\cite{peng2020convolutional}}} &
    \multirow{2}{*}{\thead{P2S\\\cite{Point2Mesh}}} &
    \multirow{2}{*}{\thead{IF\\\cite{chibane2020implicit}}} &
    \multirow{2}{*}{\thead{POCO\\\cite{boulch2022poco}}} &
    \multirow{2}{*}{\thead{NKSR\\\cite{huang2023neural}}} &
    \multicolumn{3}{c}{\thead{SurfR}}\\  \cmidrule(lr){6-8}
    &
    &
    &
    &
    \multicolumn{1}{c}{} &
    \textbf{64} &
    \textbf{128} &
    \textbf{256} \\ \midrule
    2.31 & 
    232 & 9.2 & 51 & 0.5 & 0.2 & 1.8 & 7 \\ \toprule
    \end{tabular}}}
    \label{tab:timings}
\end{table}

\paperpar{Qualitative results} 
We visualize some meshes reconstructed at a resolution of 256 in \cref{fig:qual_comp} with different noise levels for the input point cloud. Finally, we show reconstructions of two real-world objects in \cref{fig:qual_comp_real}.
Qualitatively, SurfR's results are better than most baselines, preserving more detail and completing sparsely sampled surfaces better. The reconstructed surfaces are rougher than POCO's, but the object details are still captured. The other methods over-smooth the surface, losing detail and producing holes.

\section{Limitations}\label{ss:limitations}

\begin{figure}
    \centering
    \begin{subfigure}[b]{0.36\linewidth}
        \centering
        \includegraphics[width=\textwidth]{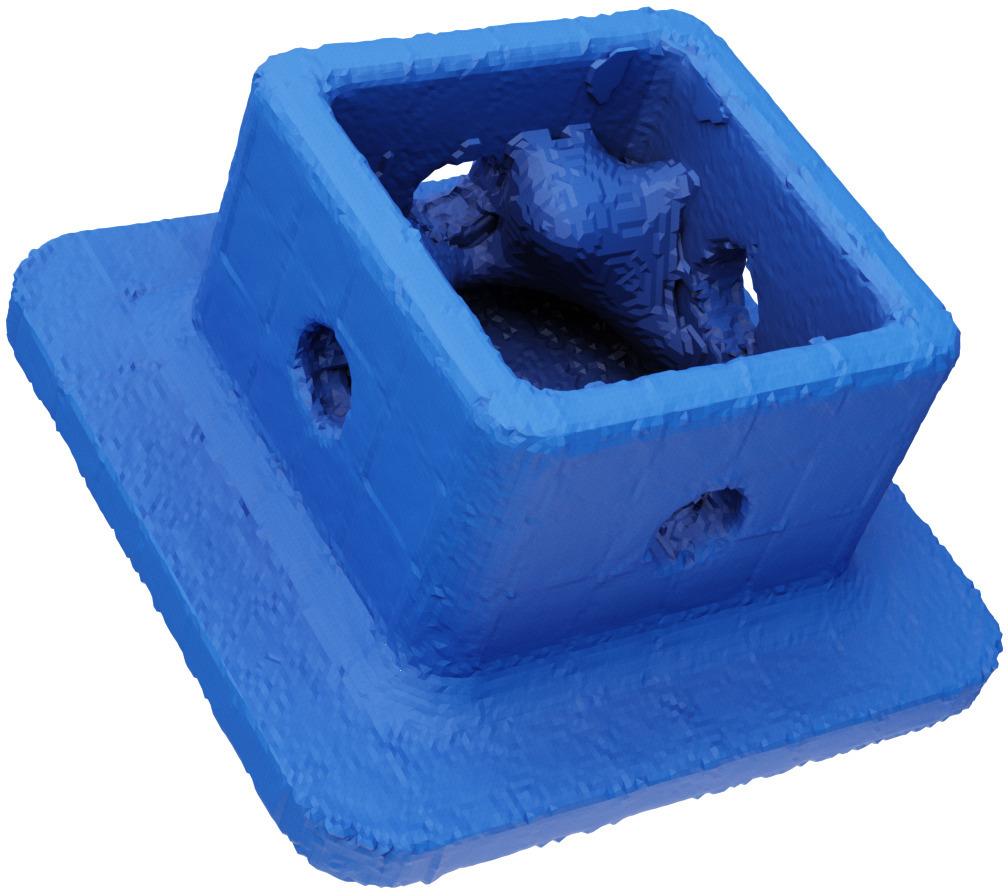}
        \vfill
        \caption{Bump-like artifacts}
        \label{subfig:bump}
    \end{subfigure}
    \quad 
    \begin{subfigure}[b]{0.36\linewidth}
        \centering
        \includegraphics[width=0.85\textwidth]{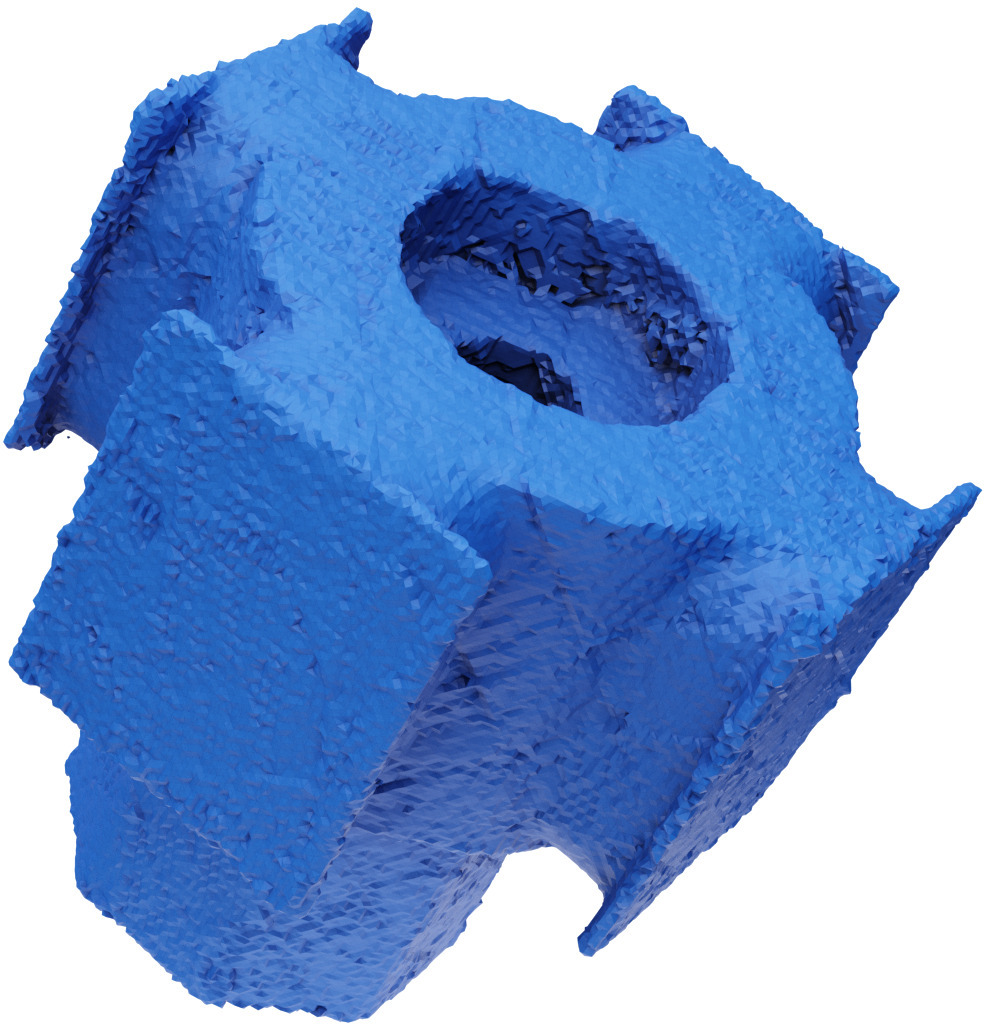}
        \vfill
        \caption{Roughness}
        \label{subfig:rough}
    \end{subfigure}
    \caption{\textbf{Limitations}: (a) shows Bump-like surfaces show due to the sign propagation during evaluation, and (b) Surface roughness in noisy input point cloud.}
    \label{fig:limitations}

\end{figure}

To obtain faster evaluation speeds, the strategy used in this work can sometimes result in the network assigning the same sign to points on both sides of the surface, leaving it up to the sign propagation step to decide the surface boundary, which results in bump-like artifacts. \Cref{fig:limitations}~\subref{subfig:bump} illustrates this limitation of the proposed method. Moreover, from the sparse nature of the architecture, rougher surfaces tend to appear when noise levels increase, as shown in~\cref{fig:limitations}~\subref{subfig:bump}. Due to the convolutional layers, other methods can smooth the surface, but at the cost of losing detail.

\section{Discussion}\label{sec:discussion}

We present SurfR, a new method for accurate and efficient implicit surface reconstruction from point clouds. 
The key to efficiency is to show a novel lazy/deferred query feature sampling technique that allows us to extract features without relying on the query points. 
Other novel contributions include techniques for using parallel multi-scale features and attention across features to improve surface reconstruction results significantly. The proposed method strikes the right balance between accuracy and speed. We demonstrate that this approach is faster than the current state-of-the-art at their optimal resolutions while achieving compelling results, preserving details, and handling sparse data.
{\bf Acknowledgements:} G. Dias Pais was supported by FCT grant {\tt PD/BD/150630/2020}.
Miraldo was supported exclusively by Mitsubishi Electric Research Laboratories.

{
\small
\bibliographystyle{ieeenat_fullname}
\bibliography{references}
}

\end{document}